\documentclass[10pt,journal,compsoc]{IEEEtran}
\ifCLASSOPTIONcompsoc
  \usepackage[nocompress]{cite}
\else
  \usepackage{cite}
\fi
\ifCLASSINFOpdf
\else
\fi

\usepackage{amsmath,amsfonts}
\usepackage{algorithmic}
\usepackage{bm,bbm}
\usepackage{color}
\usepackage{array,xcolor,colortbl}
\usepackage{booktabs}
\usepackage{multirow}
\usepackage{microtype}
\usepackage{subcaption}
\usepackage{graphicx}
\usepackage{makecell}
\usepackage{xspace}
\usepackage{enumitem}
\usepackage[breaklinks=true,
colorlinks = true,
linkcolor = red,
urlcolor  = magenta,
citecolor = blue,
anchorcolor = blue]{hyperref}
\usepackage{pifont}

\newcommand{\myparagraph}[1]{\vspace{6pt}\noindent{\bf #1}}

\definecolor{DarkBlue}{rgb}{0.0, 0.5, 0.8}
\newcommand{\yang}[1]{{ #1}}

\definecolor{VeryDarkGreen}{rgb}{0,0.5,0}

\newcommand{\revision}[1]{{ #1}}
\newcommand{\revisionR}[1]{{ #1}}

\definecolor{HL}{rgb}{0.9,0.9,0.9}
\definecolor{colorNoShape}{RGB}{153, 102, 51}
\definecolor{colorSingleModel}{RGB}{51, 153, 102}
\definecolor{colorMultiModel}{RGB}{42, 82, 162}

\newcommand{\ssp}[1]{\,{#1}\,}

\newcommand{\ie}{\textit{i.e.}\xspace}
\newcommand{\eg}{\textit{e.g.}\xspace}
\newcommand{\etal}{\textit{et al.}\xspace}

\newcommand{\dataQ}{x}
\newcommand{\annot}{y}
\newcommand{\dataC}{z}
\newcommand{\DataC}{Z}
\newcommand{\clsSet}{\DataC}
\newcommand{\DataQ}{\mathcal{X}}
\newcommand{\Annot}{\mathcal{Y}}
\newcommand{\hpz}{\hphantom{0}}
\newcommand{\hpmo}{\!\!\!}
\newcommand{\VpH}{\vphantom{$X^{X^X}$}}

\newcommand{\crop}{\text{crop}}

\newcommand{\QueryEncoder}{\mathcal{F}^\mathrm{qry}}
\newcommand{\ClassEncoder}{\mathcal{F}^\mathrm{cls}}
\newcommand{\Aggregator}{\mathcal{A}}
\newcommand{\Predictor}{\mathcal{P}}

\newcommand{\featQ}{\mathrm{f}^{\mathrm{qry}}}
\newcommand{\featC}{\mathrm{f}^{\mathrm{cls}}}
\newcommand{\featA}{\mathrm{f}^{\mathrm{agg}}}

\newcommand{\Loss}{\mathcal{L}}
\newcommand{\Cls}{C}

\newcommand{\ClsBase}{\Cls_\text{base}}
\newcommand{\ClsNovel}{\Cls_\text{novel}}
\newcommand{\Obj}{\mathsf{Obj}}

\newcommand{\predImg}{\mathsf{img}}
\newcommand{\predCls}{\mathsf{cls}}
\newcommand{\predBox}{\mathsf{box}}
\newcommand{\predMask}{\mathsf{mask}}
\newcommand{\predAng}{\mathsf{ang}}
\newcommand{\predAzi}{\mathsf{azi}}
\newcommand{\predEle}{\mathsf{ele}}
\newcommand{\predInp}{\mathsf{inp}}

\usepackage{url}
\begin{document}
\title{Few-shot Object Detection and Viewpoint Estimation for Objects in the Wild}

\author{Yang~Xiao,
        Vincent~Lepetit,
        Renaud~Marlet
\IEEEcompsocitemizethanks{\IEEEcompsocthanksitem Y. Xiao, V. Lepetit and R. Marlet are with LIGM, Ecole des Ponts, Univ Gustave Eiffel, CNRS, Marne-la-Vall\'ee, France. R.~Marlet is also with valeo.ai, Paris, France.\protect\\
E-mail: \{yang.xiao, vincent.lepetit, renaud.marlet\}@enpc.fr
}
}


\IEEEtitleabstractindextext{%
\begin{abstract}

Detecting objects and estimating their viewpoints in images are key tasks of 3D scene understanding. Recent approaches have achieved excellent results on very large benchmarks for object detection and viewpoint estimation. However, performances are still lagging behind for novel object categories with few samples. 
In this paper, we tackle the problems of few-shot object detection and few-shot viewpoint estimation. 
\revision{We demonstrate on both tasks the benefits of guiding the network prediction with class-representative features extracted from data in different modalities: image patches for object detection, and aligned 3D models for viewpoint estimation.}
Despite its simplicity, our method outperforms state-of-the-art methods by a large margin on a range of datasets, including PASCAL and COCO for few-shot object detection, and Pascal3D+ and ObjectNet3D for few-shot viewpoint estimation. 
\revision{Furthermore, when the 3D model is not available, we introduce a simple category-agnostic viewpoint estimation method by exploiting geometrical similarities and consistent pose labeling across different classes. While it moderately reduces performance, this approach still obtains better results than previous methods in this setting.}
Last, for the first time, we tackle the combination of both few-shot tasks, on three challenging benchmarks for viewpoint estimation in the wild, ObjectNet3D, Pascal3D+ and Pix3D, showing very promising results.

\end{abstract}

}

\maketitle
\IEEEdisplaynontitleabstractindextext
\IEEEpeerreviewmaketitle


\IEEEraisesectionheading{\section{Introduction}\label{sec:introduction}}

\begin{figure*}[!ht]
\centering
\includegraphics[width=0.9\linewidth]{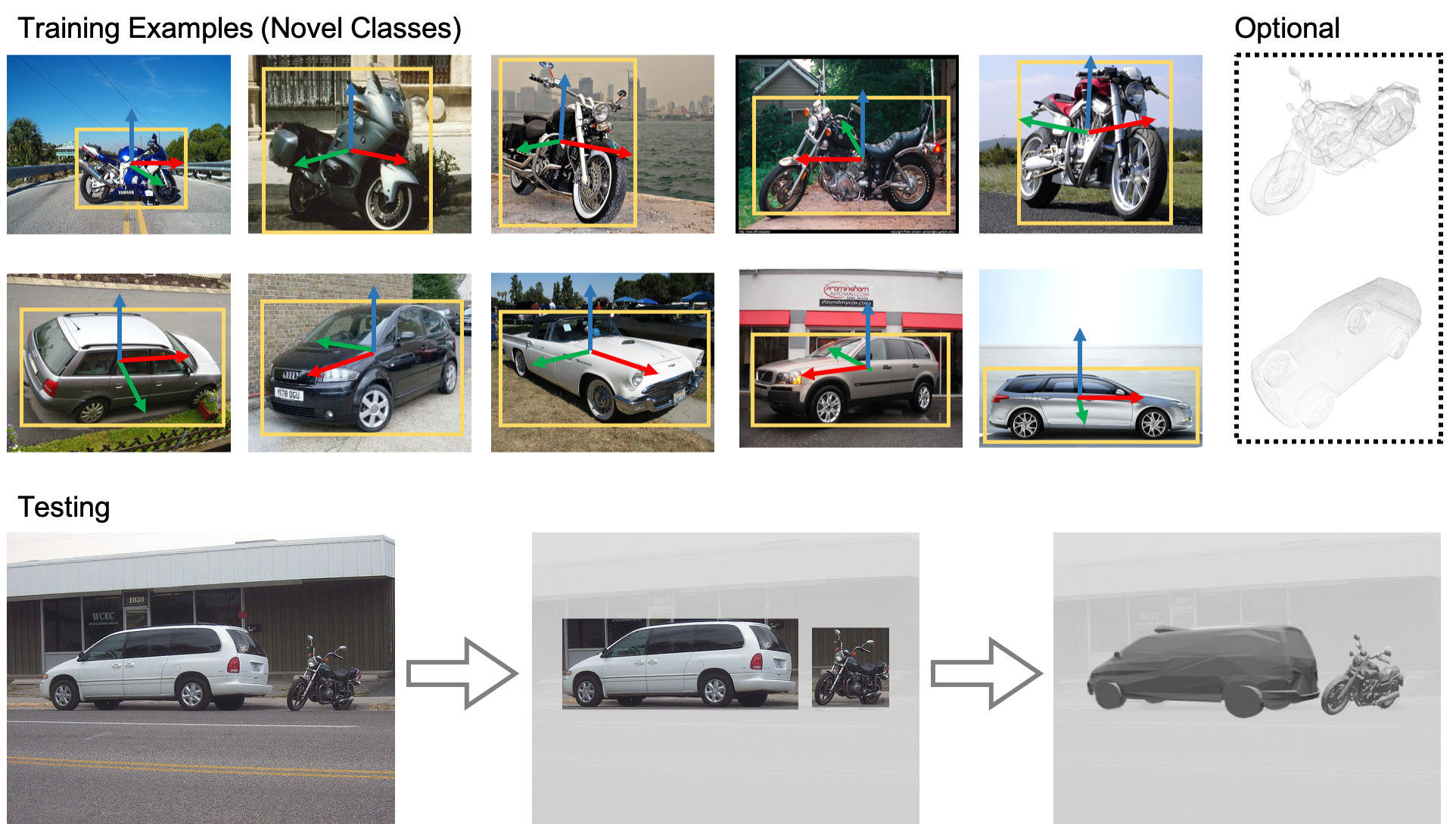}
\caption{{Few-shot object detection and viewpoint estimation.} 
Starting with images labeled with bounding boxes and viewpoints of objects from base classes, and given only a few similarly labeled images for new categories (top), we predict in a query image the 2D location of objects of new categories, as well as their 3D poses, optionally leveraging just a few arbitrary 3D class models~(bottom).
To the best of our knowledge, we are the first to conduct this joint task of object detection and viewpoint estimation in the few-shot regime.
}
\label{fig:teaser}
\end{figure*}

\IEEEPARstart{D}{etecting} objects in 2D images and estimating their 3D pose, as shown in Figure~\ref{fig:teaser}, is extremely useful for applications such as 3D scene understanding, augmented reality and robot manipulation. With the emergence of large databases annotated with object bounding boxes and viewpoints, deep-learning-based methods have achieved very good results on both tasks. However these methods, because they rely on rich labeled data, usually fail to generalize to \emph{novel} object categories when only a few annotated samples are available. Additionally, creating 3D annotations is tedious and requires a large amount of expert effort, which slows down the applications of these methods to new objects. \emph{Few-shot learning}, \ie, being able to transfer the knowledge learned from large base categories with abundant annotated images to novel categories with scarce annotated samples is therefore highly desirable in this context.

To address the few-shot learning of object detection, some approaches simultaneously tackle few-shot classification and few-shot localization by disentangling the learning of category-agnostic and category-specific network parameters~\cite{MetaDet2019}. 
\revision{Others extract a class-informative feature vector for each class and use these vectors to reweight full-image features~\cite{YOLO-FS2019} or region-of-interest (RoI) features~\cite{metarcnn2019}.
This reweighting module computes a feature similarity between query images and support classes, which has also been demonstrated to be useful in few-shot instance segmentation~\cite{Fan2020FGNFG} and few-shot image classification~\cite{Vinyals2016MatchingNF}.
However, this reweighting can easily be affected by noisy class-informative features, especially in the few-shot setting where only a few labeled samples are provided for novel categories.
Instead, we propose \revisionR{to rely on a slightly more complex combination 
of query-image features and class-informative features.
We show} that this more general aggregation module can provide better few-shot object detection performances with smaller variations when experimented with different choices of support images. Besides, it can also be used to exploit class-exemplar 3D models for few-shot viewpoint estimation.
Furthermore, we explore the usage of a cosine-similarity-based classifier~\cite{chen2019closerfewshot,wang2020few} and find that it 
slightly improves the detection results.}

In parallel to the endeavours made in few-shot object detection, recent work proposes to perform category-agnostic viewpoint estimation that can be directly applied to novel object categories without retraining~\cite{starmap2018,Xiao2019PoseFromShape}.
However, these methods either require the testing categories to be similar to the training ones~\cite{starmap2018}, or assume the exact CAD model to be provided for each object during inference~\cite{Xiao2019PoseFromShape}.
Differently, the meta-learning-based method MetaView~\cite{Tseng2019FewShotVE} introduces the category-level few-shot viewpoint estimation problem and addresses it by learning to estimate category-specific keypoints, requiring extra annotations.

While MetaView~\cite{Tseng2019FewShotVE} has achieved significantly improved performance on novel categories for few-shot viewpoint estimation, there are two main disadvantages: 1)~specific keypoints have to be designed for different object categories, which requires some expertise and can be difficult to annotate and estimate for tiny and occluded objects; 2)~the number of class-specific keypoint estimation branches increases linearly with the number of object classes.

Instead, we rely on a category-agnostic viewpoint estimation network, that \revisionR{directly predicts three Euler angles from an image embedding, without explicit class knowledge.
To that end, we exploit the fact that similar classes, e.g., \emph{sofa} and \emph{chair}, often have a consistent canonical pose, with aligned similarities. The reason probably is that many objects are consistently oriented with respect to verticality according to their regular usage. Besides, objects often present a main vertical symmetry plane and/or a notion of front and back. This is enough to define a somehow ``natural'' canonical frame, possibly up to symmetry, even for remotely-related classes such as \emph{chair} and \emph{bed}. 
Then, leveraging pose consistency across pictured classes,
we learn a category-agnostic image feature embedding space by sharing the network weights between all categories. This allows the network to exploit the geometrical similarities shared across different categories. As for the few-shot task, we first train on base classes and then simply operate a balanced fine-tuning with both novel and base classes, which is simpler and more direct than explicit feature-comparison approaches used in other class-agnostic few-shot methods \cite{zhang2019canet,yang2021classagnostic}, towards segmentation or object counting. Like for any few-shot method, our strategy works all the better when there are more similarities between base and novel classes.}
Despite its simplicity, we find that this category-agnostic prediction approach does not only outperform the state-of-the-art methods on few-shot viewpoint estimation, but also largely reduces the network complexity.

Moreover, we propose to optionally use 3D models, which we call "exemplar 3D models", as additional input of the viewpoint estimation network and to condition the final viewpoint prediction on both the image embeddings and the 3D model embeddings through a feature aggregation module. 
These 3D models are easy to obtain for many categories~\cite{shapenet2015}. They do not need to correspond exactly to the objects present in the input image---in fact we use the same exemplar 3D model for all the objects of a same category.
Their purpose is only to help the viewpoint estimation network generalize better to new classes. 
The use of these exemplar 3D models for viewpoint estimation is similar to exploiting images annotated with bounding boxes for object detection, from which we extract the task-aware class-specific information.
Using this information, we obtain an embedding for each class and condition the network prediction on both the class-informative embeddings and instance-wise query image embeddings through a feature aggregation module. This exploitation of 3D models leads to a clear performance improvement of viewpoint estimation on novel classes under the few-shot learning regime.

Finally, by combining our few-shot object detection with our few-shot viewpoint estimation, we address the joint problem of learning to detect objects in images and to estimate their viewpoints from only a few shots. This corresponds to the real world in contrast with other few-shot viewpoint estimation methods, that only evaluate in the ideal case with ground-truth classes and ground-truth bounding boxes. We demonstrate that our few-shot viewpoint estimation method can achieve very good results even based on the predicted classes and bounding boxes.

To summarize, our contributions are three-fold. 
{\bf First}, we define a simple yet effective unifying framework that addresses both few-shot object detection and few-shot viewpoint estimation in images, and achieves state-of-the-art performances across various benchmarks.
{\bf Second}, we show how the performance of our category-agnostic few-shot viewpoint estimation method is boosted by the additional knowledge at training time of one or a few exemplar 3D models per class, requiring only viewpoint supervision (as opposed to extra annotations such as keypoints), which is a realistic scenario.
{\bf Third}, we propose an evaluation of the new few-shot learning task of jointly detecting objects and estimating their viewpoint, for which we provide promising results.
Our data and code are available at \url{http://imagine.enpc.fr/~xiaoy/FSDetView/}.

This paper is an extended version of our previous work~\cite{Xiao2020FSDetView}, with several improvements:
\begin{itemize}
    \item introducing a category-agnostic few-shot viewpoint estimation method that predicts viewpoint directly from image embeddings, without relying on any 3D models during training and testing.
    \item providing a more in-depth explanation of implementation details and a thorough analysis of different components of the method.
    \item extended evaluation of joint few-shot object detection and viewpoint estimation on Pascal3D+ and Pix3D.
\end{itemize}


\section{Related work}
\label{sec:relatedWork}

Since there is a vast amount of literature on both object detection and viewpoint estimation, we focus here on recent work that targets these tasks in the case of limited labels.

\myparagraph{Few-shot learning.} 
Few-shot learning has been defined for the purpose of transferring the knowledge learned from large base categories with abundant annotated samples to novel categories with only a few annotated samples. 
Li \etal~\cite{Li2006OneshotLO} employ Bayesian inference to generalize knowledge from a pre-trained model to perform one-shot learning. 
While some methods propose to hallucinate additional training examples for the data-starved novel classes~\cite{Hariharan2016LowShotVR,Schwartz2018DeltaencoderAE,Wang2018LowShotLF,Xian2019FVAEGAND2AF}, recent work is more focused on meta-learning~\cite{Bertinetto2016LearningFO,Andrychowicz16,Vinyals2016MatchingNF,Snell2017PrototypicalNF,Hu2017RelationNF,Ravi2017OptimizationAA,lee2019meta,Hu2020Empirical}\revisionR{, which we detail below}.

Such meta-learning-based methods can be roughly divided into three categories.
1)~Metric-learning-based approaches~\cite{Vinyals2016MatchingNF,Snell2017PrototypicalNF,Hu2017RelationNF,Oreshkin2018TADAMTD,Li2019FindingTF,lee2019meta,Hou2019CrossAN,Li2019LGMNetLT} aim to learn an embedding space that is efficiently transferable for scarcely annotated training samples. MatchingNet~\cite{Vinyals2016MatchingNF} uses the cosine similarity to find the most similar class for the query image among a small set of labeled images. ProtoNet~\cite{Snell2017PrototypicalNF} replaces the weighted nearest neighbor classifier in~\cite{Vinyals2016MatchingNF} by a linear classifier where the squared Euclidean distance is used. RelationNet~\cite{Hu2017RelationNF} proposes to learn the relation between support data and query data through a neural network, which is similar to CAN~\cite{Hou2019CrossAN} and LGM-Net~\cite{Li2019LGMNetLT}.
2)~Optimization-based fast adaptation approaches~\cite{Andrychowicz16,Ravi2017OptimizationAA,Finn2017MAML,Finn2018ProbabilisticMM,Sun2019MetaTransferLF} intend to adjust the optimization algorithm such that the model can quickly converge on the few annotated samples. Ravi and Larochelle~\cite{Ravi2017OptimizationAA} train a LSTM-based meta-learner to learn a classifier in new few-shot tasks.
Model-Agnostic Meta-Learning (MAML)~\cite{Finn2017MAML} explicitly optimizes the parameters of the model such that a small number of gradient descents on the novel task will produce good generalization performance. Sun \etal~\cite{Sun2019MetaTransferLF} propose to adapt a model for few-shot learning tasks by learning scaling and shifting functions of model weights for multiple tasks.
3)~Parameter-prediction-based approaches~\cite{Bertinetto2016LearningFO,gidaris2018dynamic,Qiao2018FewShotIR} attempt to generate network parameters for new tasks. Bertinetto \etal~\cite{Bertinetto2016LearningFO} learn the parameters of factorized weight layers based on a single example of each class. Gidaris and Komodakis~\cite{gidaris2018dynamic} introduce an attention-based few-shot classification weight generator.

Besides the standard few-shot learning setting, there is also other work focused on different settings. In transductive few-shot learning~\cite{liu2019fewTPN,Hu2020Empirical}, the unlabeled query set is assumed to be accessible for training and testing. This is highly related to the semi-supervised few-shot learning~\cite{ren18fewshotssl,garcia2017few}, where an extra unlabeled training set is allowed.
These approaches only tackle the problem of few-shot image classification, while we seek to study the more challenging and under-explored problem of few-shot object detection and viewpoint estimation.

\myparagraph{Object detection with limited annotations.}
The general deep-learning models for object detection can be divided into two groups: proposal-based methods and direct methods without proposals.
While the R-CNN series~\cite{girshickICCV15fastrcnn,renNIPS15fasterrcnn,He2017MaskR} and FPN~\cite{Lin2016FeaturePN} fall into the former line of work, the YOLO series~\cite{Redmon2015YouOL,Redmon2016YOLO9000BF} and SSD~\cite{Liu2016SSDSS} belong to the latter.
All these methods mainly focus on learning from abundant data to improve detection regarding accuracy and speed. Yet, there are also some attempts to solve the problem with limited labeled data.

Chen \etal~\cite{LSTD2018} propose an approach based on transfer learning to train a network to detect objects of novel classes from just a few annotated images in the target domain.
Recent work~\cite{YOLO-FS2019,metarcnn2019} proposes to integrate a reweighting module in existing detection models such as YOLO or Faster R-CNN, which enables the network to learn generalizable features and automatically adjust them for novel class detection through a set of class-specific coefficient vectors produced from the support samples. 
Similar to the parameter-prediction-based few-shot learning methods, Wang \etal~\cite{MetaDet2019} propose to disentangle the learning of category-agnostic and category-specific components in the detection model and learn a weight-generation module to predict category-specific parameters for novel classes.
More recently, Wang \etal~\cite{wang2020few} find that a simple fine-tuning detection model can achieve impressive results on novel classes using a category-agnostic box regressor and a cosine-similarity-based box classifier. They also analyze the variance of the detection results obtained with different support samples and show the importance of averaging evaluation results over multiple experimental runs, which has been sometimes disregarded in previous work.

In contrast, we replace the feature reweighting module in~\cite{metarcnn2019} by a feature aggregation module that achieves a better detection performance under the few-shot regime. Following~\cite{MetaDet2019,metarcnn2019,wang2020few}, we also conduct multiple experiments with randomly selected support samples and report average results to prevent biases in evaluation.

\begin{figure*}[!t]
\centering
\includegraphics[width=0.8\linewidth]{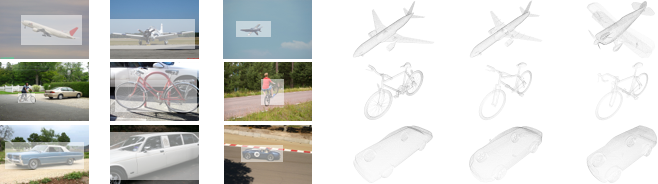} \\
\caption{Examples of class data for object detection (left) \& viewpoint estimation (right). While the images with box masks capture the characteristic appearances and the common context for different classes, the point clouds in a canonical object space capture the geometric information such as the principal axis of symmetry and the position of the main object parts.}
\label{fig:ClassData}
\end{figure*}

There is also prior work focusing on object detection with limited annotations in different settings.
Weakly-supervised detection~\cite{Song2014WeaklysupervisedDO,Bilen2015WeaklySD,Diba2016WeaklySC} considers the problem of training a detection model with only image-level labels, but without bounding box annotations that are more difficult to acquire.
Semi-supervised detection~\cite{Misra2015WatchAL,Wang2015ModelRG,Dong2019FewExampleOD} makes use of a small amount of labeled images per class to generate pseudo labels on a large amount of unlabeled images for training.
Zero-shot detection~\cite{Bansal2018ZeroShotOD,Rahman2018ZeroShotOD,Zhu2019ZeroShot} considers there is no available annotations for the novel categories and relies on external information such as inter-class relation or word embeddings for novel class detection.
Since these settings differs from the few-shot object detection setting, they are out of our scope in this work.

\myparagraph{Viewpoint estimation with limited annotations.}
Deep-learning methods for viewpoint estimation follow roughly three different paths: direct estimation of Euler angles~\cite{ViewpointsKeypoints2015,Su2015RenderFC,Mousavian20163DBB,Xiao2019PoseFromShape,Xiao2020PoseContrast}, template-based matching~\cite{Sundermeyer2018Implicit3O,Sundermeyer2020MultiPathLF} that encodes images in latent spaces and compares them against a dictionary of pre-defined viewpoints, and keypoint detection relying on 3D bounding box corners~\cite{Rad2017BB8AS,Grabner20183DPE,Pitteri2019CorNetG3} or semantic keypoints~\cite{Pavlakos20176DoFOP,starmap2018}.
Training a viewpoint estimation network requires a large amount of images manually labeled with aligned 3D CAD models or 2D keypoints, which are expensive to obtain in terms of time and human labor. To overcome this limitation, recent works propose to conduct unsupervised viewpoint estimation~\cite{Suwajanakorn2018DiscoveryOL} or predict generic 3D keypoints for all object classes~\cite{starmap2018}. Alternatively, along with the improvement of image quality and processing speed in rendering methods, abundant synthetic images can be automatically generated for network training~\cite{Su2015RenderFC,Sundermeyer2018Implicit3O,Rad2018FeatureMF,labbe2020}.
Some work also focuses on training the viewpoint estimation network on a collection of unlabeled images by self-supervised learning~\cite{deng2020self,mustikovelaCVPR20,Wang2020Self6DSM}.

Most of the existing viewpoint estimation methods are designed for known object categories or instances; very little work reports performance on unseen objects~\cite{Tulsiani2015PoseIF,starmap2018,Pitteri2019CorNetG3,Tseng2019FewShotVE,Xiao2019PoseFromShape,park2020latent}. Zhou \etal~\cite{starmap2018} propose a category-agnostic method to learn general keypoints for both seen and unseen objects, while Xiao \etal~\cite{Xiao2019PoseFromShape} show that better results can be obtained when exact 3D models of the objects are additionally provided. Park \etal~\cite{park2020latent} propose a novel framework for 6D pose estimation of unseen objects by learning a latent 3D representation from a set of reference views for each target object during inference.
In contrast to these category-agnostic methods, Tseng \etal~\cite{Tseng2019FewShotVE} specifically address the few-shot scenario by training a category-specific viewpoint estimation network for novel classes with limited samples.
\revision{More recently, Wang \etal~\cite{wang2021neural} study the problem of learning to estimate the 3D object pose from a few labeled examples and a collection of unlabeled data, and show promising results in particular on vehicle categories.}

Instead of using exact 3D object models as~\cite{Xiao2019PoseFromShape}, we propose a meta-learning approach to extract a class-informative canonical shape feature vector for each novel class from a few labeled samples, with random object models.
Besides, our network can be applied to both base and novel classes without changing the network architecture, while~\cite{Tseng2019FewShotVE} requires a separate meta-training procedure for each class and needs keypoint annotations in addition to the viewpoint.


\section{Method}
\label{sec:method}

In this section, we first introduce the setup for few-shot object detection and few-shot viewpoint estimation (Section~\ref{sec:fsSetup}).
Then, we present our network architecture for these two tasks with class data (Section~\ref{sec:network}) and a fine-tuning category-agnostic viewpoint estimation method (Section~\ref{sec:fsViewAgnostic}).
Finally, we describe the learning procedure adopted in both few-shot learning tasks (Section~\ref{sec:LearningProc}).

\begin{figure*}[!t]
    \centering
    \includegraphics[width=0.9\linewidth]{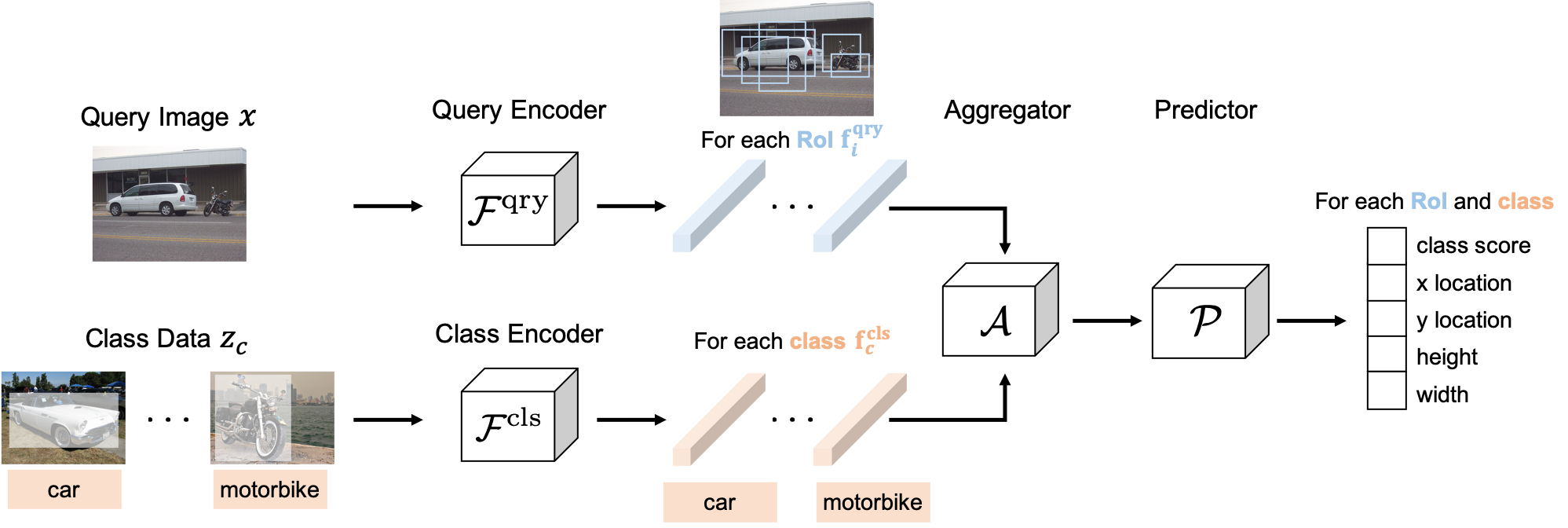} \\
    (a) {\bf few-shot object detection.} \\ 
    \vspace{3mm}
    \includegraphics[width=0.9\linewidth]{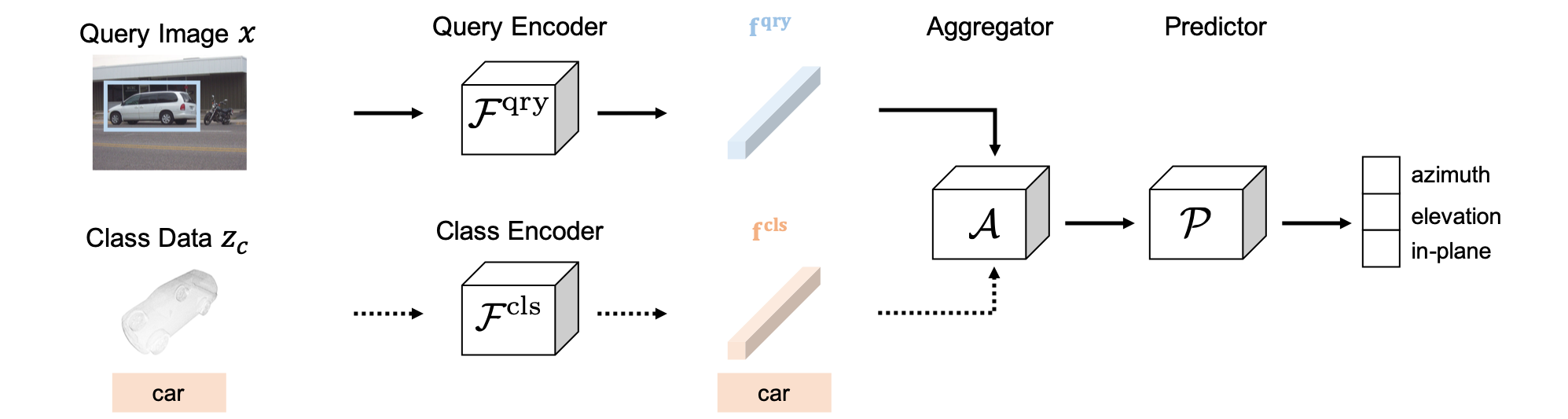} \\
    (b) {\bf few-shot viewpoint estimation.} \\
    \caption{
    \textbf{Method overview.}
    \textbf{(a)} For object detection, we sample for each class~$c$ one image $\dataQ$ in the training set containing an object $j$ of class~$c$, to which we add an extra channel for the binary mask $\predMask_j$ of the ground-truth bounding box $\predBox_j$ of object $j$. Each corresponding vector of class features $\featC_c$ (red) is then combined with each vector of query features $\featQ_i$ (blue) associated to one of the region of interest $i$ in the query image, via an aggregation module. Finally, the aggregated features $\featA_{i,c}$ pass through a predictor that estimates a class probability $\predCls_{i,c}$ and regresses a bounding box $\predBox_{i,c}$.
    \textbf{(b)} For few-shot viewpoint estimation, we represent the 3D pose using three Euler angles. We estimate them either directly from the query features extracted from the image or, optionally, indirectly from aggregated features made of both query features and class information extracted from a few point clouds with coordinates in a normalized, canonical object space.
    }
    \label{fig:overview}
\end{figure*}

\subsection{Few-shot Learning Setup}
\label{sec:fsSetup}

For both the object detection and viewpoint estimation tasks, we assume we have training samples $(\dataQ, \annot)\ssp\in(\DataQ, \Annot)$. A few 3D shapes may also be available for viewpoint estimation.
\begin{itemize}
    \item In the case of object detection, $\dataQ$ is an image, $\annot \,{=}\, \{(\predCls_i,\predBox_i) \,{\mid}\, i\in\Obj_\dataQ\}$ indicates the class label $\predCls_i$ and bounding box $\predBox_i$ of each object~$i$ in the image.
    
    \item In the case of viewpoint estimation, $\dataQ \,{=}\, (\predCls,\predBox,\predImg)$ represents an object of class $\predCls(\dataQ)$ pictured in bounding box $\predBox(\dataQ)$ of an image $\predImg(\dataQ)$; $\annot = \predAng = (\predAzi, \predEle, \predInp)$ is the 3D pose (viewpoint) of the object, given by Euler angles (azimuth, elevation, in-plane rotation).
\end{itemize}
For each class $c \ssp\in C \ssp= \{\predCls_i \ssp\mid \dataQ\ssp\in\DataQ, i\ssp\in\Obj_\dataQ\}$, we consider a set $\DataC_c$ of \emph{class data} (see Figure~\ref{fig:ClassData}) to learn from using meta-learning:
\begin{itemize}
    \item For object detection, $\DataC_c \ssp= \{(\dataQ,\predMask_i)\ssp\mid \dataQ\ssp\in\DataQ, i\ssp\in\Obj_\dataQ\}$ is made of images $\dataQ$ plus an extra channel with a binary mask for bounding box $\predBox_i$ of object $i\ssp\in\Obj_\dataQ$.
    
    \item For viewpoint estimation, $\DataC_c$ is an optional, additional set of 3D models of class $c$, which is not used in the purely image-based category-agnostic variant.
\end{itemize}    
At each training iteration, class data $\dataC_c$ is randomly sampled in $\clsSet_c$ for each $c\ssp\in C$.

In the few-shot setting, we have a partition of the classes $\Cls = \ClsBase \,\cup\, \ClsNovel$, with many samples for base classes in $\ClsBase$ and only a few samples (possibly also including a few shapes) for novel classes in $\ClsNovel$.  The goal is to transfer the knowledge learned on base classes with abundant samples to little-represented novel classes.

\subsection{Few-shot Learning with Class Data}
\label{sec:network}

Our general approach has three steps illustrated in Figure~\ref{fig:overview}.
First, query data $\dataQ$ and class-informative data $\dataC_c$ pass respectively through the query encoder $\QueryEncoder$ and the class encoder $\ClassEncoder$ to generate corresponding feature vectors\revisionR{, for each each region of interest~(RoI) and each class respectively}.
Next, a feature aggregation module $\Aggregator$ 
\revisionR{combines a query feature (for a given RoI) with a class feature.}
Finally, the output of the network is obtained by passing \revisionR{each} 
aggregated feature 
through a task-specific predictor $\Predictor$:
\begin{itemize}
    \item For object detection, the predictor estimates a classification score and an object location (i.e.., bounding box) for each region of interest~(RoI) and each class.
    \item For viewpoint estimation, the predictor selects quantized angles by classification, that are refined using regressed angular offsets.
\end{itemize}

\begin{figure*}[!t]
\centering
\includegraphics[width=1.0\linewidth]{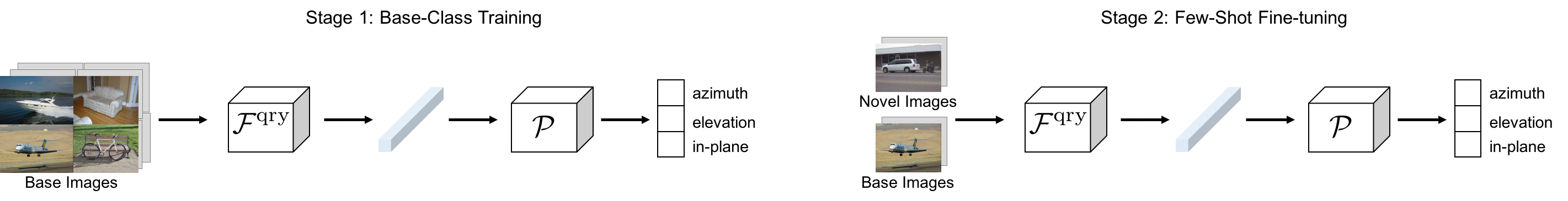} \\
\caption{Illustration of our category-agnostic viewpoint estimation approach without using 3D models. The network is first trained on abundant labeled images of base classes (left), then fine-tuned on a balanced set of images containing both base and novel classes (right).}
\label{fig:TrainStages}
\end{figure*}

\subsubsection{Few-shot Object Detection.}
We adopt the popular Faster R-CNN~\cite{renNIPS15fasterrcnn} approach in our few-shot object detection network (see Figure~\ref{fig:overview}(a)).
The query encoder $\QueryEncoder$ includes the backbone, the region proposal network (RPN) and the proposal-level feature alignment module. 
\yang{In parallel, the class encoder $\ClassEncoder$ is the backbone sharing the same weights as $\QueryEncoder$ except for the first convolutional layer, that has an additional fourth channel for extracting class features from RGB images with binary masks of the object bounding boxes~\cite{YOLO-FS2019,metarcnn2019}.}
Each extracted vector of query features is aggregated with each extracted vector of class features before being processed for class classification and bounding box regression:
\begin{equation} \label{eq:DetPred}
\begin{aligned}
    &( \predCls_{i, c}, \predBox_{i, c} ) = \Predictor \Big(\Aggregator\big(\featQ_i, \featC_c \big) \Big)
    \\
    &\quad\text{for~~}
    \featQ_i\ssp\in\QueryEncoder(\dataQ),\ 
    \featC_c = \ClassEncoder(\dataC_c),\ 
    c \in \Cls_\mathrm{train}
\end{aligned}
\end{equation}
where $\Cls_\mathrm{train}$ is the set of all training classes, and where $\predCls_{i,c}$ and $\predBox_{i,c}$ are the predicted classification scores and object locations for the $i^\mathrm{th}$ RoI in query image $\dataQ$ and for class $c$.
The prediction branch $\Predictor$ is \revision{implemented as two fully-connected layers of size 4096 without activation that output respectively $N_\mathrm{train}\ssp=|\Cls_\mathrm{train}|$ classification scores and $N_\mathrm{train}$ box regressions for each RoI.} 
The final predictions are obtained by concatenating all the class-wise network outputs.

\myparagraph{Cosine similarity for box classifier.}
Inspired by Wang \etal \cite{wang2020few}, we use a cosine-similarity-based classifier in the bounding box predictor.
We note the weight matrix of the box classifier as $\mathrm{W} = [\mathrm{w}_1, \mathrm{w}_2, \dots, \mathrm{w}_c]$, where $\mathrm{w}_c \in \mathbb{R}^d$ is the class-wise weight vector and $d$ is the dimension of the aggregated features.
Thus, the classification score for the $i^\mathrm{th}$ RoI and class $c$ can be written as:
\begin{equation} \label{eq:CosineCls}
    \predCls_{i,c} = \frac{\alpha \Aggregator\big(\featQ_i, \featC_c \big)^\top \mathrm{w}_c}{\| \Aggregator\big(\featQ_i, \featC_c \big) \| \| \mathrm{w}_c \|} \> ,
\end{equation}
where $\alpha$ is a scaling factor, set to 20 in all experiments.
The instance-level feature normalization used in this cosine-similarity-based classifier was found empirically to be helpful in reducing the intra-class variance and improving the detection accuracy of novel classes.

\subsubsection{Few-shot Viewpoint Estimation.}\label{sec:fsviewestim}

For few-shot viewpoint estimation, we rely on the recently proposed PoseFromShape~\cite{Xiao2019PoseFromShape} architecture to implement our network.
To create class data $\dataC_c$, we transform the 3D models in the dataset into point clouds by uniformly sampling points on the surface, with coordinates in a normalized, canonical object space. 
The query encoder $\QueryEncoder$ and class encoder $\ClassEncoder$ (cf.\ Figure~\ref{fig:overview}(b)) correspond respectively to the image encoder ResNet-18~\cite{He2015ResNet} and shape encoder PointNet~\cite{Qi2016PointNetDL} in PoseFromShape.
By aggregating the query features and class features, we estimate the three Euler angles \revision{via the predictor $\Predictor$, which is implemented as a three-layer fully-connected network of sizes 800, 400, 200, each layer being followed by a batch normalization and ReLU activation}:
\begin{equation} \label{eq:ViewPred}
\begin{aligned}
    (\predAzi, \predEle, \predInp) &= \Predictor \Big(\Aggregator\big(\featQ, \featC \big) \Big)
    \\
    \quad\text{with~~}
    \featQ &= \QueryEncoder(\crop(\predImg(\dataQ), \predBox(\dataQ)))\text{, and}\\
    \featC &= \ClassEncoder(\dataC_c),\ 
    c = \predCls(\dataQ)
\end{aligned}
\end{equation}
where $\crop(\predImg(\dataQ), \predBox(\dataQ))$ indicates that the query features are extracted from the object-centred crops.
Unlike the object detection making a prediction for each class, 
here we only make the prediction for the object class $\predCls(\dataQ)$ by passing the corresponding class data through the network.
We also use the mixed classification-and-regression viewpoint estimator of~\cite{Xiao2019PoseFromShape}: the output consists of angular bin classification scores and within-bin offsets for three Euler angles: azimuth ($\predAzi$), elevation ($\predEle$), and in-plane rotation ($\predInp$).

\subsubsection{Feature Aggregation.}
In recent few-shot object detection methods such as FSRW~\cite{YOLO-FS2019} and Meta R-CNN~\cite{metarcnn2019}, features are aggregated by reweighting the query features $\featQ$ according to the output $\featC$ of the class encoder $\ClassEncoder$:
\begin{equation} \label{eq:featRW}
\Aggregator(\featQ, \featC) = \featQ \odot \featC \> ,
\end{equation}
where $\odot$ represents \revision{element-wise multiplication (Hadamard product)} and $\featQ$ has the same number of channels as $\featC$. 
By jointly training the query encoder $\QueryEncoder$ and the class encoder $\ClassEncoder$ with this reweighting module, it is possible to learn to generate meaningful reweighting vectors~$\featC$. $\QueryEncoder$ and $\ClassEncoder$ actually share their weights, except the first layer~\cite{metarcnn2019}.

We choose to rely on a slightly more complex aggregation scheme. The fact is that feature subtraction is a different but also effective way to measure similarity between image features~\cite{ammiratoTDID18,Kuo2019ShapeMaskLT}. The image embedding $\featQ$ itself, without any reweighting, contains relevant information too. Our aggregation thus concatenates the three forms: 
\begin{equation} \label{eq:featAgg}
\Aggregator(\featQ, \featC) = [\featQ \odot \featC, \featQ - \featC, \featQ] \> ,
\end{equation}
where $[\cdot, \cdot, \cdot]$ represents channel-wise concatenation. The last part of the aggregated features in Eq.~\eqref{eq:featAgg} is independent of the class data. 
\yang{As observed experimentally in Table~\ref{tab:DetAblation}, this partial disentanglement does not only improve few-shot detection performance, it also reduces the variation introduced by the randomness of support samples.}

\subsection{Category-agnostic Viewpoint Estimation}
\label{sec:fsViewAgnostic}

We also consider the case where no 3D model is provided. In this case, we bypass the requirement of task-aware class data as mentioned in the previous section and we estimate viewpoints only from the image embeddings.
Given a query object $\dataQ$ pictured in image $\predImg(\dataQ)$ and its bounding box $\predBox(\dataQ)$, the query encoder generates an image embedding $\featQ$. 
Then, given such an embedding, the viewpoint prediction component estimates the three Euler angles:
\begin{equation} \label{eq:predBaseline}
\begin{aligned}
    (\predAzi, \predEle, \predInp) &= \Predictor \Big(\featQ\Big)
    \\
    \quad\text{with~~}
    \featQ &= \QueryEncoder(\crop(\predImg(\dataQ), \predBox(\dataQ))) \> .
\end{aligned}
\end{equation}
The feature extraction module is category-agnostic and all object classes share the same prediction module. 
\revision{And the viewpoint predictor $\Predictor$ is implemented in the same way as in Section~\ref{sec:fsviewestim}.}
Therefore, the network can fully leverage the geometrical similarities between related categories such as \emph{bicycle} and \emph{motorbike}.

As illustrated in Figure~\ref{fig:TrainStages}, we first train on a large base-class dataset and then fine-tune on a balanced dataset consisting of base and novel classes. \revisionR{While not exactly following the general framework of Figure~\ref{fig:overview}, it follows a
related pattern, where the 3D branch is removed, as well as, consequently, the aggregation module.} This simple yet effective approach outperforms previous methods on few-shot viewpoint estimation (see Section~\ref{sec:expView}).

\revisionR{Following previous few-shot approaches, we fine-tune the network on both base and novel categories for ``learning without forgetting'', which prevents the network to only focus on increasing its performance on novel categories ignoring possible dramatic drops on base categories.}

\subsection{Learning Procedure}
\label{sec:LearningProc}

Our learning procedure consists of two phases: \emph{base-class training} on many samples from base classes ($\Cls_{\mathrm{train}}=\Cls_{\mathrm{base}}$), followed by \emph{few-shot fine-tuning} on a balanced small set of samples from both base and novel classes ($\Cls_{\mathrm{train}}=\Cls_{\mathrm{base}} \cup \Cls_{\mathrm{novel}}$). 
\revision{More precisely, in the $K$-shot fine-tuning stage where only $K$ labeled samples are available for each novel class, we randomly select $K$ samples for each base class to balance the training iterations between base and novel classes.
}
In both phases, we optimize the network using the same loss function.

\subsubsection{Loss Function}

\myparagraph{Detection loss function.}

We optimize our few-shot object detection network using the same loss function as Meta R-CNN~\cite{metarcnn2019}:
\begin{equation} \label{eq:lossDet}
    \Loss = \Loss_\mathrm{rpn} + \Loss_\mathrm{cls} + \Loss_\mathrm{loc} + \Loss_\mathrm{meta} \> ,
\end{equation}
where $\Loss_\mathrm{rpn}$ is applied to the output of the RPN to distinguish foreground from background and refine the proposals, $\Loss_\mathrm{cls}$ is a cross-entropy loss for box classification, $\Loss_\mathrm{loc}$ is a Huber loss for box regression, and $\Loss_\mathrm{meta}$ is a cross-entropy loss encouraging class features to be diverse for different classes~\cite{metarcnn2019}.

\myparagraph{Viewpoint loss function.}
For the task of estimating viewpoints, we discretize each Euler angle with a bin size of 15 degrees and use the same loss function as PoseFromShape~\cite{Xiao2019PoseFromShape} to train the network:
\begin{equation} \label{eq:lossView}
    \Loss = \sum_{\theta \in \{\predAzi, \predEle, \predInp\}} \Loss_\mathrm{cls}^\theta + \Loss_\mathrm{reg}^\theta \> ,
\end{equation}
where $\Loss_\mathrm{cls}^\theta$ is a cross-entropy loss for angle bin classification of Euler angle $\theta$, and $\Loss_\mathrm{reg}^\theta$ is a Huber loss for the regression of offsets relatively to bin centers.
Here we remove the meta loss $\Loss_\mathrm{meta}$ used in object detection since we want the network to learn useful inter-class similarities for viewpoint estimation, instead of the inter-class differences for box classification in object detection.

\subsubsection{Class Data Construction}
For viewpoint estimation, unless otherwise stated, we make use of all the 3D models available for each class (typically less than 10) during both training stages.
In contrast, the class data used in object detection requires the information of object class and location, which is limited for novel classes by the number of annotated samples.
Therefore, we use a large number of class data for base classes in the base training stage (typically $|\clsSet_c| = 200$, as in Meta R-CNN~\cite{metarcnn2019}) and limit the size of $\clsSet_c$ to the number of shots for both base and novel classes in the $K$-shot fine-tuning stage ($|\clsSet_c| = K$).

For inference, instead of randomly sampling class data from the dataset as done during training, we construct class features once and for all after learning is finished: for each class~$c$, we average all class features used in the few-shot fine-tuning stage:
\begin{equation} \label{eq:inferFeatMeta}
\featC_c = \frac{1}{|\clsSet_c|} \sum_{\dataC_c \in \clsSet_c} \ClassEncoder(\dataC_c) \> .
\end{equation}
This corresponds to the offline computation of all \revisionR{orange} feature vectors in Figure~\ref{fig:overview}(a).


\section{Experiments}
\label{sec:exp}

In this section, we first evaluate on few-shot object detection (Section~\ref{sec:expDet}) and few-shot viewpoint estimation benchmarks (Section~\ref{sec:expView}) to empirically assess the effectiveness of our method.
For a fair comparison, we use the same splits between base and novel classes as used in previous work~\cite{YOLO-FS2019,Tseng2019FewShotVE} and report the performance averaged over multiple runs with different groups of few-shot training examples to obtain a sensible accuracy estimation~\cite{Tseng2019FewShotVE,wang2020few}.
Furthermore, we conduct an evaluation of the joint task of few-shot object detection and viewpoint estimation on three datasets to demonstrate the generalization capacity of our method for both tasks in the few-shot regime (Section~\ref{sec:expJoint}). We conclude this empirical study with limitations of our approach (Section~\ref{sec:limitations}).

\subsection{Few-shot Object Detection}
\label{sec:expDet}

We adopt a well-established evaluation protocol for few-shot object detection~\cite{YOLO-FS2019,MetaDet2019,metarcnn2019} and report performance on PASCAL VOC~\cite{PascalVOC10,PascalVOC15} (reported in Table~\ref{tab:DetVOC}) and MS-COCO~\cite{Lin2014MicrosoftCOCO} (reported in Table~\ref{tab:DetCOCO}).

\subsubsection{Experimental Setup}

\myparagraph{Datasets.}
PASCAL VOC~\cite{PascalVOC10,PascalVOC15} is a small-scale object detection dataset containing 20 object categories. 
Following the common protocol~\cite{Redmon2016YOLO9000BF,girshickICCV15fastrcnn,renNIPS15fasterrcnn}, we use the test set of VOC 2007~\cite{PascalVOC10} for testing and the train-val set of VOC 07-12~\cite{PascalVOC15} for training, which results in 16,551 training images and 4,952 testing images.
Among the 20 object categories, \cite{YOLO-FS2019} introduces three few-shot splits by randomly selecting 5 classes as the novel ones while keeping the remaining 15 ones as the base: 
({\it bird, bus, cow, motorbike, sofa / rest}); ({\it aeroplane, bottle, cow, horse, sofa / rest}); ({\it boat, cat, motorbike, sheep, sofa / rest}).
We evaluate on these 3 different base/novel splits assuming that only $K$ annotated bounding boxes are provided for each novel class during training, where $K$ equals 1, 2, 3, 5 or 10.

MS-COCO~\cite{Lin2014MicrosoftCOCO} is a large-scale object detection dataset containing 80 object categories.
We follow~\cite{YOLO-FS2019,metarcnn2019,MetaDet2019} to use 5,000 images from the mini-val set for testing and use the remaining 118,287 images in train-val set for training.
Among the 80 object categories, we select the 20 classes common to PASCAL VOC as novel classes and consider the remaining 60 classes as base classes.
For this dataset, the evaluation protocol used in previous work is to test on $K = \mathrm{10}$ or 30 annotated bounding boxes for each novel class.

\begin{table*}[!t]
\addtolength{\tabcolsep}{4pt}
\begin{center}
\caption{{\bf Few-shot object detection evaluation on PASCAL VOC.} We report the Average Precision with a single IoU threshold at 0.5 ($\mathrm{AP}^{0.5}$) under 3 different splits for 5 novel classes \cite{YOLO-FS2019} with a small number of shots. *Results computed over single experimental run with a fix set of support images.
}
\label{tab:DetVOC}
	\scalebox{0.85}{
	\begin{tabular}{l | ccccc | ccccc | ccccc}
	\toprule
	& \multicolumn{5}{c|}{Novel Set 1} & \multicolumn{5}{c|}{Novel Set 2} & \multicolumn{5}{c}{Novel Set 3} \\
	Method \textbackslash\ Shots & 1 & 2 & 3 & 5 & 10 & 1 & 2 & 3 & 5 & 10 & 1 & 2 & 3 & 5 & 10 \\
	\midrule
	LSTD~~\cite{LSTD2018}* & \hpz8.2  & \hpz1.0  & 12.4 & 29.1 & 38.5 & 11.4 & \hpz3.8  & \hpz5.0  & 15.7 & 31.0 & 12.6 & \hpz8.5  & 15.0 & 27.3 & 36.3 \\
	FSRW~~\cite{YOLO-FS2019}* & 14.8 & 15.5 & 26.7 & 33.9 & 47.2 & 15.7 & 15.2 & 22.7 & 30.1 & 40.5 & {21.3} & 25.6 & 28.4 & 42.8 & 45.9 \\
	MetaDet~~\cite{MetaDet2019} & 18.9 & 20.6 & 30.2 & 36.8 & 49.6 & {\bf 21.8} & 23.1 & 27.8 & 31.7 & 43.0 & 20.6 & 23.9 & 29.4 & 43.9  & 44.1 \\
	Meta R-CNN~~\cite{metarcnn2019} & 19.9 & 25.5 & 35.0 & 45.7 & 51.5 & 10.4 & 19.4 & 29.6 & 34.8 & {45.4} & 14.3 & 18.2 & 27.5 & 41.2 & 48.1 \\
	TFA w/fc~~\cite{wang2020few} & 22.9 & 34.5 & 40.4 & 46.7 & 52.0 & 16.9 & 26.4 & 30.5 & 34.6 & 39.7 & 15.7 & 27.2 & 34.7 & 40.8 & 44.6 \\
	TFA w/cos~~\cite{wang2020few} & {25.3} & {\bf 36.4} & 42.1 & 47.9 & 52.8 & 18.3 & {\bf 27.5} & {30.9} & 34.1 & 39.5 & 17.9 & 27.2 & 34.3 & 40.8 & 45.6 \\
	\hline
	Ours w/fc \VpH & {24.2} & {35.3} & {42.2} & {\bf 49.1} & {57.4} & {21.6} & {24.6} & {\bf 31.9} & {37.0} & {\bf 45.7} & {21.2} & {30.0} & {\bf 37.2} & {\bf 43.8} & {49.6} \\
	Ours w/cos & {\bf 26.9} & {35.7} & {\bf 42.3} & {48.9} & {\bf 57.8} & {21.2} & {26.7} & {30.6} & {\bf 37.7} & {45.1} & {\bf 24.3} & {\bf 30.4} & {36.3} & {41.6} & {\bf 50.1} \\
	\bottomrule
	\end{tabular}}
\end{center}
\end{table*}

\begin{table*}[!t]
\begin{center}
\addtolength{\tabcolsep}{6pt}
\caption{{\bf Few-shot object detection evaluation on MS-COCO.} We report the standard MS-COCO evaluation metrics on the 20 novel classes of COCO. *Results computed over single experimental run with a fix set of support images.
}
\label{tab:DetCOCO}
    \scalebox{0.85}{
	\begin{tabular}{cl |ccc ccc |ccc ccc}
	\toprule
	& & \multicolumn{6}{c|}{Average Precision} & \multicolumn{6}{c}{Average Recall} \\
	Shots & Method & 0.5:0.95 & 0.5 & 0.75 & S & M & L & 1 & 10 & 100 & S & M & L \\
	\midrule
	\multirow{10}{*}{10} & LSTD~~\cite{LSTD2018}* & 3.2 & \hpz8.1 & \hpz2.1 & 0.9 & \hpz2.0 & \hpz6.5 & \hpz7.8 & 10.4 & 10.4 & 1.1 & \hpz5.6 & 19.6 \\
	& FSRW~~\cite{YOLO-FS2019}* & 5.6 & 12.3 & \hpz4.6 & 0.9 & \hpz3.5 & 10.5 & 10.1 & 14.3 & 14.4 & 1.5 & \hpz8.4 & 28.2 \\
	& MetaDet~~\cite{MetaDet2019} & 7.1 & 14.6 & \hpz6.1 & 1.0 & \hpz4.1 & 12.2 & 11.9 & 15.1 & 15.5 & 1.7 & \hpz9.7 & 30.1 \\
	& Meta R-CNN~~\cite{metarcnn2019} & 8.7 & 19.1 & \hpz6.6 & 2.3 & \hpz7.7 & 14.0 & 12.6 & 17.8 & 17.9 & 7.8 & 15.6 & 27.2 \\
	& FSOD~~\cite{fan2020fsod}* & \hpmo11.1 & 20.4 & 10.6 & -- & \hpz-- & -- & -- & -- & -- & -- & -- & -- \\
	& MPSR~~\cite{wu2020mpsr}* & 9.8 & 17.9 & \hpz9.7 & {\bf 3.3} & \hpz9.2 & 16.1 & 15.7 & 21.2 & 21.2 & 4.6 & 19.6 & 34.3 \\
	& TFA w/fc~~\cite{wang2020few} & 9.1 & 17.3 & \hpz8.5 & -- & \hpz-- & -- & -- & -- & -- & -- & -- & -- \\
	& TFA w/cos~~\cite{wang2020few} & 9.1 & 17.1 & \hpz8.8 & -- & \hpz-- & -- & -- & -- & -- & -- & -- & -- \\ 
	\cline{2-14}
	& Ours w/fc \VpH & {12.5} & {27.3} & {\hpz9.8} & {2.5} & {13.8} & {19.9} & {20.0} & {25.5} & {25.7} & {7.5} & {27.6} & {38.9} \\
	& Ours w/cos & {\bf 13.6} & {\bf 28.6} & {\bf 11.3} & {2.6} & {\bf 14.6} & {\bf 22.1} & {\bf 20.6} & {\bf 26.8} & {\bf 27.0} & {\bf 7.9} & {\bf 28.8} & {\bf 41.3} \\
	\midrule
	
	\multirow{9}{*}{30} & LSTD~~\cite{LSTD2018}* & \hpz6.7 & 15.8 & \hpz5.1 & 0.4 & \hpz2.9 & 12.3 & 10.9 & 14.3 & 14.3 & 0.9 & \hpz7.1 & 27.0 \\
	& FSRW~~\cite{YOLO-FS2019}* & \hpz9.1 & 19.0 & \hpz7.6 & 0.8 & \hpz4.9 & 16.8 & 13.2 & 17.7 & 17.8 & 1.5 & 10.4 & 33.5 \\
	& MetaDet~~\cite{MetaDet2019} & 11.3 & 21.7 & \hpz8.1 & 1.1 & \hpz6.2 & 17.3 & 14.5 & 18.9 & 19.2 & 1.8 & 11.1 & 34.4 \\
	& Meta R-CNN~~\cite{metarcnn2019} & 12.4 & 25.3 & 10.8 & 2.8 & 11.6 & 19.0 & 15.0 & 21.4 & 21.7 & 8.6 & 20.0 & 32.1 \\
	& MPSR~~\cite{wu2020mpsr}* & 14.1 & 25.4 & 14.2 & {\bf 4.0} & 12.9 & 23.0 & 17.7 & 24.2 & 24.3 & 5.5 & 21.0 & 39.3 \\
	& TFA w/fc~~\cite{wang2020few} & 12.0 & 22.2 & 11.8 & -- & -- & -- & -- & -- & -- & -- & -- & -- \\
	& TFA w/cos~~\cite{wang2020few} & 12.1 & 22.0 & 12.0 & -- & -- & -- & -- & -- & -- & -- & -- & -- \\
	\cline{2-14}
	& Ours w/fc \VpH & {14.7} & {30.6} & {12.2} & {3.2} & {15.2} & {23.8} & {22.0} & {28.2} & {28.4} & {8.3} & {30.3} & {42.1} \\
	& Ours w/cos & {\bf 16.4} & {\bf 32.6} & {\bf 14.7} & {3.5} & {\bf 17.1} & {\bf 26.2} & {\bf 23.3} & {\bf 29.7} & {\bf 29.9} & {\bf 8.8} & {\bf 31.9} & {\bf 44.7} \\
	\bottomrule
	\end{tabular}}
	\end{center}
\end{table*}

\myparagraph{Evaluation metrics.}
We measure the Average Precision (AP) of detections as the area under a precision-recall curve.
For few-shot object detection on PASCAL VOC, we classically report $\mathrm{AP}^{0.5}$, that computes AP with a single Intersection over Union~(IoU) threshold at 0.5. 
For evaluation on MS-COCO, we use the standard MS-COCO evaluation metrics~\cite{Redmon2016YOLO9000BF,renNIPS15fasterrcnn}: mAP, $\mathrm{AP}^{0.5}$, $\mathrm{AP}^{0.75}$, $\mathrm{AP^{S}}$, $\mathrm{AP^{M}}$, $\mathrm{AP^{L}}$, $\mathrm{AR}^{1}$, $\mathrm{AR}^{10}$, $\mathrm{AR}^{100}$, $\mathrm{AR^{S}}$, $\mathrm{AR^{M}}$, $\mathrm{AR^{L}}$.
While $\mathrm{AP}^{0.5}$ and $\mathrm{AP}^{0.75}$ represent respectively the AP with a single IoU threshold at 0.5 and 0.75, mAP is the averaged AP over multiple IoU thresholds from 0.5 to 0.95 with a step of 0.05. 
Average Recall~(AR) computed with the $N$ most confident predictions per image is noted as $\mathrm{AR}^N$, where $N$ equals 1, 10 or 100.
Moreover, we report the detection performance across different object scales: S (small: $\mathrm{area} < 32^2$ square pixels), M (medium: $32^2 \leq \mathrm{area} < 96^2$) and L (large: $96^2 \leq \mathrm{area}$).

\myparagraph{Training details.}
We employ the same learning scheme as~\cite{metarcnn2019}, which uses the SGD optimizer with an initial learning rate of $10^{-3}$ and a batch size of 4. Weight decay and momentum are set to 0.0005 and 0.9, respectively.
In the first training stage, we train for 20 epochs and divide the learning rate by 10 after each 5 epochs. In the second stage, we train for 5 epochs with a learning rate of $10^{-3}$ and another 4 epochs with a learning rate of $10^{-4}$.
For anchor scales, we use three scales ($128^2, 256^2, 512^2$) for PASCAL VOC and add a fourth scale of $64^2$ for MS-COCO. The three aspect ratios of anchors are set to 1:2, 1:1, 2:1.
We augment the data with horizontal flipping.
Training on a single Titan-X GPU takes around one day for PASCAL VOC and ten days for MS-COCO.

\begin{table*}[t]
\addtolength{\tabcolsep}{6pt}
    \centering
    \caption{{\bf Ablation study on the feature aggregation scheme.}
    Using the same class splits of PASCAL VOC as in Table~\ref{tab:DetVOC}, we measure the performance of few-shot object detection on the novel classes for 3 shots and 10 shots. We report the average and standard deviation of the AP50 metric over ten runs. $\featQ$ is the query features and $\featC$ is the class features.}
    \vspace{-2mm}
    \label{tab:DetAblation}
    \scalebox{0.9}{
    \begin{tabular}{l  c c c c c c}
    \toprule
    & \multicolumn{2}{c}{Novel Set 1} & \multicolumn{2}{c}{Novel Set 2} & \multicolumn{2}{c}{Novel Set 3}\\
    Method $\backslash$ \ Shots & 3 & 10 & 3 & 10 & 3 & 10 \\
    \midrule
    $[ \featQ \odot \featC ]$ & $35.0 \pm 3.6$ & $51.5 \pm 5.8$ & $29.6 \pm 3.5$ & $45.4 \pm 5.5$ & $27.5 \pm 5.2$ & $48.1 \pm 5.9$ \\
    
    $[ \featQ \odot \featC , \featQ ]$ & $36.6 \pm 7.1$ & $49.6 \pm 4.3$ & $27.5 \pm 5.7$ & $41.6 \pm 3.7$ & $28.7 \pm 5.9$ & $44.0 \pm 2.7$ \\
    
    $[ \featQ \odot \featC, \featQ, \featC ]$ & $37.6 \pm 7.2$ & $54.2 \pm 4.9$ & $30.0 \pm 2.9$ & $41.0 \pm 5.3$ & $33.6 \pm 5.0$ & $47.5 \pm 2.3$ \\
    
    $[ \featQ \odot \featC, \featQ - \featC ]$ & $39.2 \pm 4.5$ & $55.5 \pm 3.9$ & $31.7 \pm 6.2$ & $45.2 \pm 3.3$ & $35.6 \pm 5.6$ & $48.9 \pm 3.3$ \\
    
    $[ \featQ \odot \featC, \featQ - \featC, \featQ ]$ & $\bm{42.2} \pm \bm{2.1}$ & $\bm{57.4} \pm \bm{2.7}$ & $\bm{31.9} \pm \bm{2.7}$ & $\bm{45.7} \pm \bm{1.8}$ & $\bm{37.2} \pm \bm{3.5}$ & $\bm{49.6} \pm \bm{2.2}$ \\
    \bottomrule
    \end{tabular}}
\end{table*}

\subsubsection{Few-shot Detection Results}

\myparagraph{Cosine similarity vs. dot product.}
We first compare the cosine-similarity-based box classifier ({Ours w/cos}) with the normal FC-based classifier ({Ours w/fc}) that uses a simple dot product between feature representations and weight vectors to compute the classification scores.
Indeed, in few-shot learning tasks, features learned with a cosine-similarity-based classifier have been found empirically to generalize better to novel categories compared to features learned with FC-based classifier~\cite{gidaris2018dynamic,wang2020few}.
As observed in Table~\ref{tab:DetVOC} and Table~\ref{tab:DetCOCO}, even though the improvement is not systematic on novel classes of PASCAL VOC, cosine similarity does bring a consistent performance boost on novel classes of COCO, compared to FC-based classifier with direct dot product. 

\myparagraph{Different feature aggregations.}
We analyze the impact of different feature aggregation schemes. 
For this purpose, we evaluate $K$-shot object detection on PASCAL VOC with $K\,{=}\,3$ or $10$. Here, we compare results obtained by models with an FC-based classifier. The results are reported in Table~\ref{tab:DetAblation}. We can see that our feature aggregation scheme $[ \featQ \odot \featC, \featQ \,{-}\, \featC, \featQ ]$ yields the best precision. In particular, although the difference $\featQ \,{-}\, \featC$ could in theory be learned from the individual feature vectors $[\featQ, \featC]$, the network performs better when explicitly provided with their subtraction. Moreover, our aggregation scheme significantly reduces the variance introduced by the random sampling of few-shot support data, which is a major issues in few-shot learning (although sometimes neglected).

\begin{table*}[!t]
\addtolength{\tabcolsep}{4pt}
\centering
	\caption{{Intra-dataset 10-shot viewpoint estimation evaluation.} 
	We report Acc30($\uparrow$) / MedErr($\downarrow$) on the same 20 novel classes of ObjectNet3D for each method, while 80 are used as base classes. All models are trained and tested on ObjectNet3D.}
	\vspace{-2mm}
	\label{tab:ViewIntra}
	\scalebox{0.9}{
	\begin{tabular}{l cccccc c}
	\toprule
	Method & bed & bookshelf & calculator & cellphone & computer & door & f\_cabinet \\ \midrule
	StarMap+F~~\cite{starmap2018} & 0.32 / 47.2 & 0.61 / 21.0 & 0.26 / 50.6 & 0.56 / 26.8 & 0.59 / 24.4 & - / - & 0.76 / 17.1 \\
	StarMap+M~~\cite{starmap2018} & 0.32 / 42.2 & 0.76 / 15.7 & 0.58 / 26.8 & 0.59 / 22.2 & 0.69 / 19.2 & - / - & 0.76 / 15.5 \\
	MetaView~~\cite{Tseng2019FewShotVE} & 0.36 / 37.5 & 0.76 / 17.2 & {\bf 0.92} / 12.3 & 0.58 / 25.1 & 0.70 / 22.2 & - / - & 0.66 / 22.9 \\
	\hline
	Ours w/o 3D \VpH & {0.53} / {26.8} & {0.82} / {\hpz9.4} & {0.76} / {11.6} & {0.54} / {24.0} & {0.82} / {11.8} & {0.86} / {3.1} & {0.83} / {11.1} \\
	Ours w/ 3D & {\bf 0.64} / {\bf 14.8} & {\bf 0.90} / {\bf \hpz7.8} & {0.90} / {\bf \hpz8.2} & {\bf 0.61} / {\bf 13.2} & {\bf 0.86} / {\bf 10.3} & {\bf 0.90} / {\bf 0.8} & {\bf 0.86} / {\bf 10.2} \\
	\midrule \midrule
	Method & guitar & iron & knife & microwave & pen & pot & rifle \\ \midrule
	StarMap+F~~\cite{starmap2018} & 0.54 / 27.9 & 0.00 / 128 & 0.05 / 120 & 0.82 / 19.0 & - / - & 0.51 / 29.9 & 0.02 / 100 \\
	StarMap+M~~\cite{starmap2018} & 0.59 / 21.5 & 0.00 / 136 & 0.08 / 117 & 0.82 / 17.3 & - / - & 0.51 / 28.2 & 0.01 / 100 \\
	MetaView~~\cite{Tseng2019FewShotVE} & {0.63} / {24.0} & 0.20 / {\hpz77} 
	& 0.05 / \textbf{\hpz98} 
	& 0.77 / 17.9 & - / - & 0.49 / 31.6 & 0.21 / {\bf \hpz81} 
	\\
	\hline
	Ours w/o 3D \VpH & {0.60} / {21.5} & {0.08} / {118} & {0.21} / {137} & {0.91} / {\hpz8.9} & {0.39} / {63.2} & {0.64} / {17.5} & {0.15} / {\hpz91} 
	\\
	Ours w/ 3D & {\bf 0.68} / {\bf 19.4} & {\bf 0.34} / {\bf \hpz60} 
	& {\bf 0.27} / {137} & {\bf 0.93} / {\bf \hpz7.4} & {\bf 0.47} / {\bf 36.4} & {\bf 0.76} / {\bf 11.8} & {\bf 0.28} / {\hpz87} 
	\\
	\midrule \midrule
	Method & shoe & slipper & stove & toilet & tub & wheelchair & \cellcolor{HL} All \\ 
	\midrule
	StarMap+F~~\cite{starmap2018} & - / - & 0.08 / 128 & 0.80 / 16.1 & 0.38 / 36.8 & 0.35 / 39.8 & 0.18 / 80.4 & \cellcolor{HL}0.41 / 41.0 \\
	StarMap+M~~\cite{starmap2018} & - / - & 0.15 / 128 & 0.83 / 15.6 & 0.39 / 35.5 & 0.41 / 38.5 & 0.24 / 71.5 & \cellcolor{HL}0.46 / 33.9 \\
	MetaView~~\cite{Tseng2019FewShotVE} & - / - & 0.07 / 115 & 0.74 / 21.7 & 0.50 / 32.0 & 0.29 / 46.5 & 0.27 / {\bf 55.8} & \cellcolor{HL}0.48 / 31.5 \\
	\hline
	Ours w/o 3D \VpH & {0.35} / {47.2} & {0.19} / {125} & {0.86} / {11.3} & {0.49} / {30.2} & {0.50} / {32.0} & {\bf 0.36} / {57.8} &\cellcolor{HL} {0.56} / {22.0} \\
	Ours w/ 3D & {\bf 0.49} / {\bf 30.6} & {\bf 0.28} / {\bf \hpz93} 
	& {\bf 0.91} / {\bf \hpz9.5} & {\bf 0.69} / {\bf 17.8} & {\bf 0.65} / {\bf 16.4} & {0.35} / {61.2} &\cellcolor{HL} {\bf 0.65} / {\bf 15.6} \\ 
	\bottomrule
	\end{tabular}}
	\vspace{3mm}
	\caption{{Inter-dataset 10-shot viewpoint estimation evaluation.}
	We report Acc30($\uparrow$) / MedErr($\downarrow$) on the 12 novel classes of Pascal3D+, while the 88 base classes are in ObjectNet3D. All models are trained on ObjectNet3D and tested on Pascal3D+.}
	\vspace{-2mm}
	\label{tab:ViewInter}
	\scalebox{0.9}{
	\begin{tabular}{l ccccccc}
	\toprule
	Method & aero & bike & boat & bottle & bus & car & chair \\
	\midrule
	StarMap+F~~\cite{starmap2018} & 0.03 / 102 & 0.05 / 98.8 & 0.07 / \hpz99 
	& 0.48 / 31.9 & 0.46 / 33.0 & 0.18 / 80.8 & 0.22 / {\bf 74.6} \\
	StarMap+M~~\cite{starmap2018} & 0.03 / \hpz99 
	& 0.08 / 88.4 & 0.11 / \hpz92 
	& 0.55 / 28.0 & 0.49 / 31.0 & 0.21 / 81.4 & 0.21 / 80.2 \\
	MetaView~~\cite{Tseng2019FewShotVE} & 0.12 / 104 & 0.08 / 91.3 & 0.09 / 108 & 0.71 / 24.0 & 0.64 / 22.8 & 0.22 / 73.3 & 0.20 / 89.1 \\
	\hline
	Ours w/o 3D \VpH & {0.14} / {\hpz88} 
	& {0.30} / {67.8} & {0.20} / {\hpz83} 
	& {0.81} / {12.1} & {0.73} / {\hpz9.6} & {0.43} / {53.8} & {0.30} / {78.8} \\
	Ours w/ 3D & {\bf 0.21} / {\bf \hpz73} 
	& {\bf 0.33} / {\bf 64.7} & {\bf 0.25} / {\bf \hpz78} 
	& {\bf 0.91} / {\bf 11.6} & {\bf 0.74} / {\bf \hpz9.0} & {\bf 0.49} / {\bf 32.8} & {\bf 0.32} / {79.1} \\
	\midrule \midrule
	Method & table & mbike & sofa & train & tv & \multicolumn{2}{c}{\cellcolor{HL} All} \\ 
	\midrule
	StarMap+F~~\cite{starmap2018} & 0.46 / 31.4 & 0.09 / 91.6 & 0.32 / 44.7 & 0.36 / 41.7 & 0.52 / 29.1 & \multicolumn{2}{c}{\cellcolor{HL} 0.25 / 64.7} \\
	StarMap+M~~\cite{starmap2018} & 0.29 / 36.8 & 0.11 / 83.5 & 0.44 / 42.9 & 0.42 / 33.9 & 0.64 / 25.3 & \multicolumn{2}{c}{\cellcolor{HL} 0.28 / 60.5} \\
	MetaView~~\cite{Tseng2019FewShotVE} & 0.39 / 36.0 & 0.14 / 74.7 & 0.29 / 46.2 & 0.61 / 23.8 & 0.58 / 26.3 & \multicolumn{2}{c}{\cellcolor{HL} 0.33 / 51.3} \\
	\hline
	Ours w/o 3D \VpH & {0.51} / {31.2} & {0.36} / {49.8} & {0.49} / {34.6} & {0.62} / {16.1} & {0.77} / {\bf 18.7} & \multicolumn{2}{c}{\cellcolor{HL} 0.46 / 38.3} \\
	Ours w/ 3D & {\bf 0.59} / {\bf 20.9} & {\bf 0.44} / {\bf 37.2} & {\bf 0.58} / {\bf 23.9} & {\bf 0.72} / {\bf 12.1} & {\bf 0.79} / {19.0} & \multicolumn{2}{c}{\cellcolor{HL} \bf {0.51} / {\bf 29.1}} \\
	\bottomrule
	\end{tabular}}
	\vspace*{-2mm}
\end{table*}

\myparagraph{Comparison with the state of the art.}
Tables~\ref{tab:DetVOC} and~\ref{tab:DetCOCO} show the comparison with previous few-shot object detection methods.
On the PASCAL VOC dataset, our method achieves the best performance in most cases, in particular when the number of shots tends to be large.
This indicates that our method can better leverage the task-relevant information from novel classes when more labeled examples are provided.
Moreover, it significantly improves results on MS-COCO for all evaluation metrics, which validates again the effectiveness of our approach.

\subsection{Few-shot Viewpoint Estimation}
\label{sec:expView}

Following the few-shot viewpoint estimation protocol proposed in~\cite{Tseng2019FewShotVE}, we evaluate our method in two settings: \emph{intra}-dataset on ObjectNet3D~\cite{objectnet3d16} (cf.\ Table~\ref{tab:ViewIntra}) and \emph{inter}-dataset between ObjectNet3D and Pascal3D+~\cite{pascal3d14} (cf.\ Table~\ref{tab:ViewInter}).  

\subsubsection{Experimental Setup}

\myparagraph{Datasets.}
Pascal3D+~\cite{pascal3d14} is a standard evaluation benchmark used in 3D pose estimation. Unlike 6D pose estimation datasets~\cite{Hinterstoier2012ModelBT,Hodan2017TLESSAR,xiang2018posecnn} 
that usually focus on dozens of objects with limited environment variations, Pascal3D+ contains 12 man-made object categories with 2k to 4k images per category,
allowing the benchmarking of object pose estimation in the wild.
ObjectNet3D~\cite{objectnet3d16}, extended from Pascal3D+, features 100 object categories, with 90,127 images and 201,888 objects in total.
In both datasets, only a small number of roughly-aligned 3D models are provided for each category.

\myparagraph{Evaluation metrics.}
We use the most common metrics for evaluation: {Acc30}, which is the percentage of estimations with a rotational error smaller than $30^\circ$, and {MedErr}, which is the median rotational error measured in degrees. 
We compute the rotational error as $\Delta (R_\mathrm{pred}, R_\mathrm{gt}) = \frac{\|\log (R_\mathrm{pred}^\top, R_\mathrm{gt})\|_F}  {\sqrt{2}}$, where $\|\cdot\|_F$ is the Frobenius norm.
Following previous work~\cite{starmap2018,Tseng2019FewShotVE}, we only use the non-occluded and non-truncated objects for evaluation, and assume in this subsection, for all methods, that the ground-truth classes and ground-truth bounding boxes are provided at test time.

\myparagraph{Training details.}
We resize the object image crops into $224\times224$ pixels as the input for our viewpoint estimation networks, with (Ours w/ 3D) or without (Ours w/o 3D) using exemplar 3D models.
Both networks are trained using the Adam optimizer with a batch size of 16. Weight decay is set to 0.0005. During the base-class training stage, we train for 150 epochs with a learning rate of $10^{-4}$. For few-shot fine-tuning, we train for 50 epochs with learning rate of $10^{-4}$ and another 50 epochs with a learning rate of $10^{-5}$.
Standard data augmentation is applied during training, such as random rotation, random flipping and color jittering.
\yang{The training is done in about one day on a single Titan-X GPU.}

\myparagraph{Compared methods.}
For few-shot viewpoint estimation, we compare our method to MetaView~\cite{Tseng2019FewShotVE} and to two adaptations of StarMap~\cite{starmap2018}.
More precisely, the authors of MetaView~\cite{Tseng2019FewShotVE} re-implemented StarMap with one stage of ResNet-18 as the backbone, and trained the network with MAML~\cite{Finn2017MAML} for a fair comparison in the few-shot regime (StarMap+M).
They also provided StarMap results by just fine-tuning it on the novel classes using the scarce labeled data (StarMap+F). 
We consider the two variants of our method, with (Ours w/ 3D) or without 3D data (Ours w/o 3D) at training time.

\subsubsection{Few-shot Viewpoint Estimation Results}

\myparagraph{Intra-dataset evaluation.}
We follow the protocol of~\cite{Tseng2019FewShotVE,Xiao2019PoseFromShape} to split the 100 categories of ObjectNet3D into 80 base classes and 20 novel classes.
As shown in Table~\ref{tab:ViewIntra}, our model outperforms the recently proposed meta-learning-based method MetaView~\cite{Tseng2019FewShotVE} by a very large margin in overall performance: $+16$ points in Acc30 and half MedErr (from $31.5^\circ$ down to $15.6^\circ$).
Besides, keypoint annotations are not available for some object categories such as door, pen and shoe in ObjectNet3D. This lack of annotations limits the generalization of keypoint-based approaches~\cite{starmap2018,Tseng2019FewShotVE} as they require a set of manually labeled keypoints for network training.
In contrast, our model can be trained and evaluated on all object classes of ObjectNet3D as we only rely on the viewpoint annotations. 
More importantly, our model can be directly deployed on different classes using the same architecture, while MetaView learns a set of separate category-specific semantic keypoint detectors for each class.
This flexibility suggests that our approach is likely to exploit the similarities between different categories (e.g., bicycle and motorbike) and has more potentials for applications to robotics and augmented reality.

\myparagraph{Inter-dataset evaluation.}
To further evaluate our method in a more practical scenario, we use a source dataset for base classes and another target dataset for novel (disjoint) classes.
Following the same split as MetaView~\cite{Tseng2019FewShotVE}, we use all 12 categories of Pascal3D+ as novel categories and the remaining 88 categories of ObjectNet3D as base categories.
Distinct from the previous intra-dataset experiment that focuses more on the cross-category generalization capacity, this inter-dataset setup also reveals the cross-domain generalization ability.

As shown in Table~\ref{tab:ViewInter}, our approach again significantly outperforms StarMap and MetaView.
Our overall improvement in inter-dataset evaluation is even larger than in intra-dataset evaluation: we gain $+19$ points in Acc30 and again divide MedErr by about 2 (from $51.3^\circ$ down to $28.3^\circ$).
This indicates that our approach, by leveraging viewpoint-relevant 3D information, not only helps the network generalize to novel classes from the same domain, but also addresses the domain shift issues when trained and evaluated on different datasets.

\begin{figure*}[p]
\centering
\includegraphics[width=1.0\linewidth]{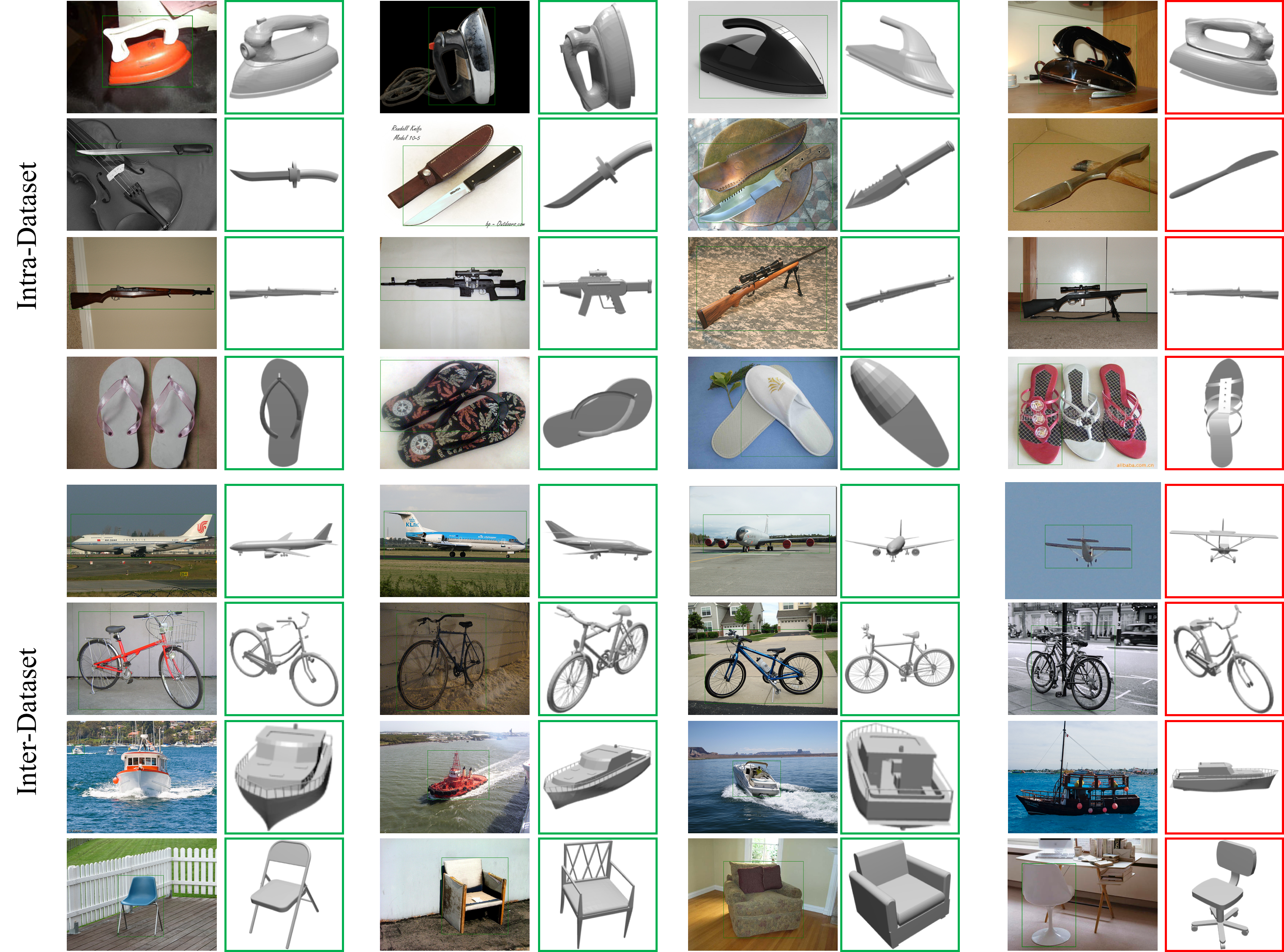}
\vspace{-4mm}
\caption{{Qualitative results of few-shot viewpoint estimation using ground-truth 2D bounding boxes (and classes).}
We visualize results on ObjectNet3D and Pascal3D+. For each category, we show three success cases (first six columns) and one failure case (last two columns). CAD models are shown here only for the purpose of illustrating the estimated viewpoint. Failure cases usually result from appearance ambiguities of a same object in different poses, or from heavily cluttered scenes.
}
\label{fig:fsView}
\vspace{3mm}
\begin{tabular}{@{}c@{\hspace*{.01\linewidth}}c@{\hspace*{.05\linewidth}}c@{\hspace*{.01\linewidth}}c@{}}
    Intra-dataset (base) & Intra-dataset (novel) & Inter-dataset (base) & Inter-dataset (novel) \\
    \includegraphics[width=.25\linewidth, height=.18\linewidth]{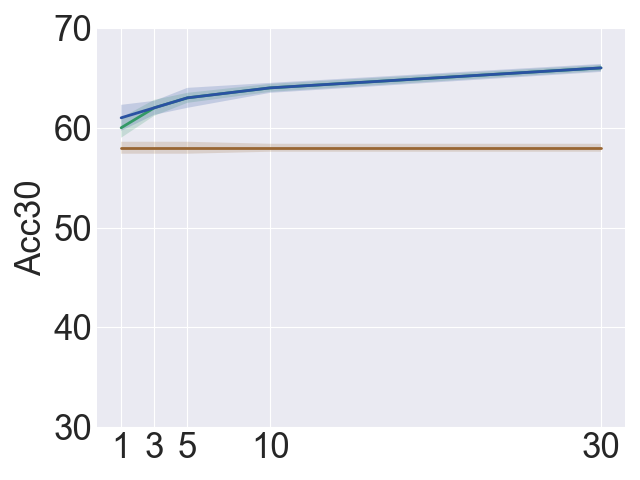}
    &
    \includegraphics[width=.22\linewidth, height=.18\linewidth]{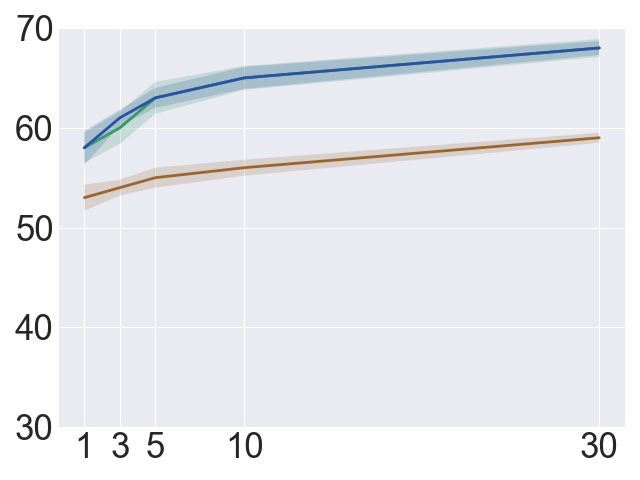}
    &
    \includegraphics[width=.22\linewidth, height=.18\linewidth]{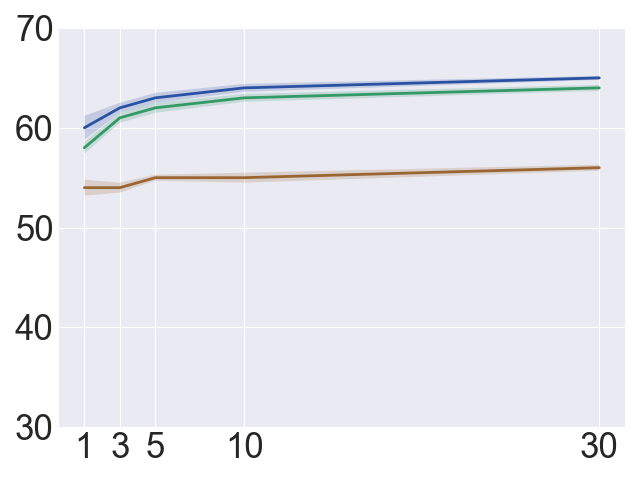}
    &
    \includegraphics[width=.22\linewidth, height=.18\linewidth]{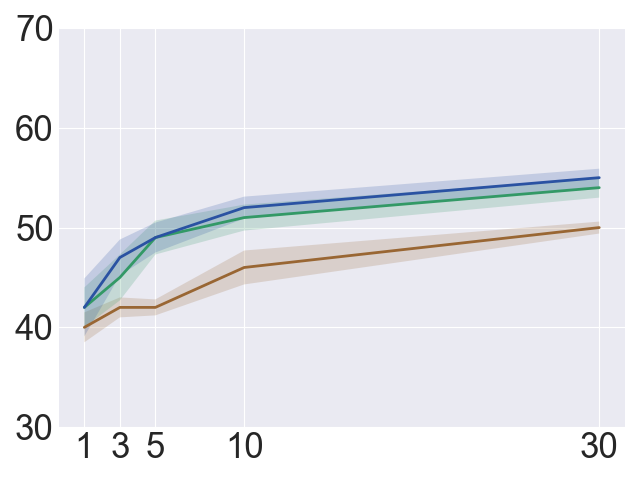} \\
    
    \includegraphics[width=.25\linewidth, height=.18\linewidth]{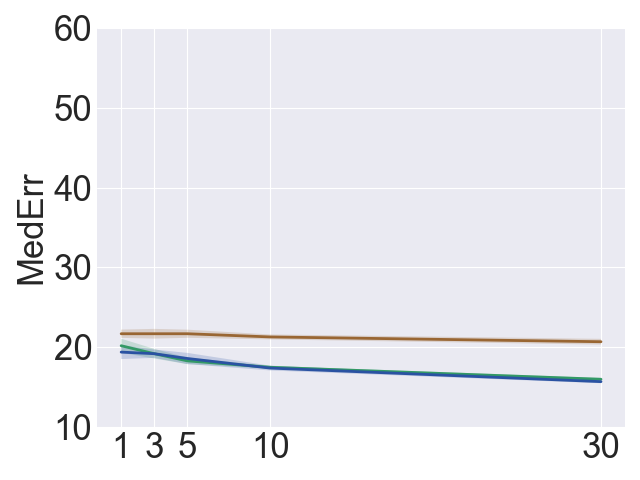}
    &
    \includegraphics[width=.22\linewidth, height=.18\linewidth]{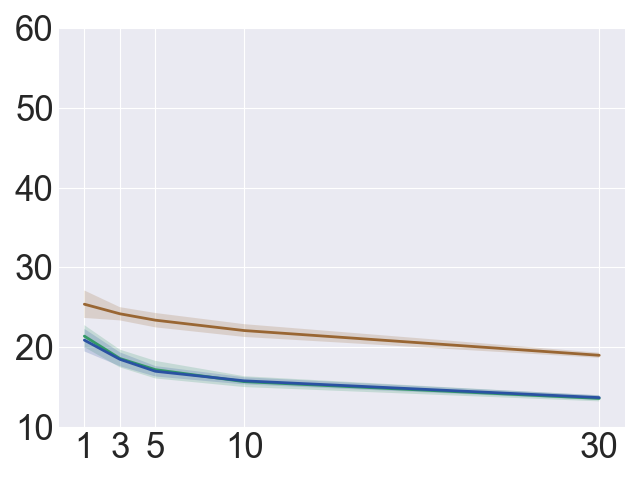}
    &
    \includegraphics[width=.22\linewidth, height=.18\linewidth]{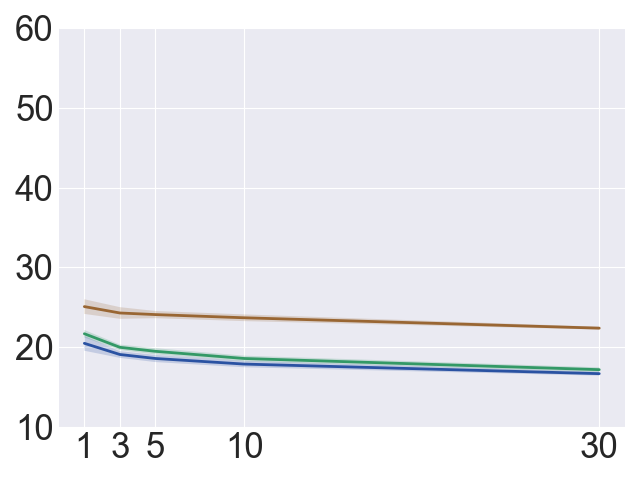}
    &
    \includegraphics[width=.22\linewidth, height=.18\linewidth]{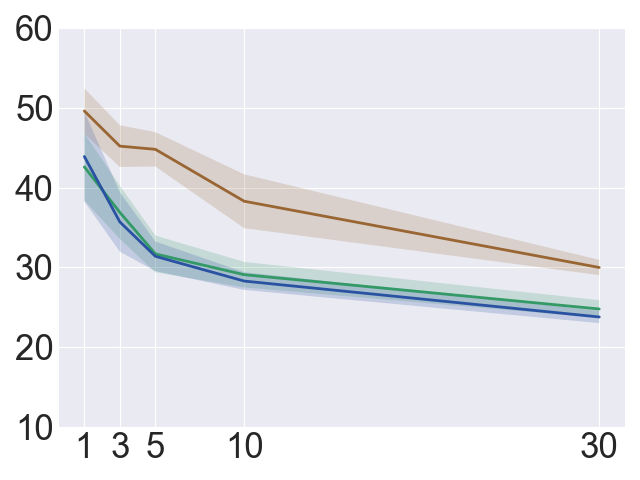} \\
\end{tabular} \\
{\small \textcolor{colorNoShape}{No 3D Exemplar} \quad \quad \textcolor{colorSingleModel}{Single 3D Exemplar} \quad \quad \textcolor{colorMultiModel}{Multiple 3D Exemplars}}
\vspace{-1mm}
\caption{{Few-shot viewpoint estimation evaluation using different number of shots.} For each metric, we report the average and standard deviation computed over 10 random experiments.}
\label{fig:ViewVaryShot}
\end{figure*}

\myparagraph{Visual results.}
We illustrate on Figure~\ref{fig:fsView} viewpoint estimation for novel objects in ObjectNet3D and Pascal3D+.
We show both success (green boxes) and failure cases (red boxes) to help analyze possible error types.
We visualize categories giving large rotational errors: \emph{iron}, \emph{knife}, \emph{rifle} and \emph{slipper} for ObjectNet3D, \emph{aeroplane}, \emph{bicycle}, \emph{boat} and \emph{chair} for Pascal3D+.
The most common failure cases come from objects with similar appearances in different poses, \eg, \emph{iron} and \emph{knife} in ObjectNet3D, \emph{aeroplane} and \emph{boat} in Pascal3D+. \revisionR{It seems that more complex methods based on keypoints \cite{Tseng2019FewShotVE, starmap2018} perform a bit better on this kind of objects, although being nevertheless grossly wrong too.}
Other failure cases include heavy clutter cases (\emph{bicycle}) and large shape variations between training objects and testing objects (\emph{chair}).

\subsubsection{Ablation Study}

\myparagraph{Different 3D model representations, if any.}
In Table~\ref{tab:Ablation3D}, we analyze the impact of different 3D model representations in our few-shot viewpoint estimation approach using exemplar 3D models.
Besides using a point cloud ({Point Cloud}), we can also represent 3D shapes using a group of depth images ({Depth}) or non-textured rendered images ({Rendering}) captured in a set of camera locations defined on the upper hemisphere. 
We also use the normalized, canonical object space~\cite{Brachmann2014Learning6O,Wang_2019_NOCS,grabner2019LFD} to represent the 3D models by transforming the 3D coordinates into RGB values ({Object Coord.}).
For these variants that consider 2D inputs rather than a 3D point cloud, we implement the class encoder $\ClassEncoder$ using a ResNet-18 to extract features from images.

\begin{table}[!t!]
\centering
\caption{Efficacy of different 3D representations, if any. We show few-shot viewpoint estimation results on the 20 novel classes of ObjectNet3D. The first row represents our approach without using any form of 3D information, while other rows correspond to our method using exemplar 3D models with different representations.
We also plot the four different 3D representations of an example CAD model on the bottom.}
\label{tab:Ablation3D}
    \scalebox{0.9}{
    \begin{tabular}{l|c c}
    \toprule
    & \multicolumn{2}{c}{{Acc30}($\uparrow$) / {MedErr}($\downarrow$)} \\
    3D exemplar & Base & Novel \\
    \midrule
    None & $0.58\pm0.01$ / $21.3\pm0.31$ & $0.56\pm0.01$ / $22.1\pm0.80$ \\
    \midrule
    Depth & $0.61\pm0.01$ / $22.0\pm0.97$ & $0.57\pm0.02$ / $24.3\pm1.52$ \\
    Object Coord. & $0.61\pm0.01$ / $22.0\pm0.54$ & $0.59\pm0.02$ / $23.7\pm1.09$ \\
    Rendering & $0.61\pm0.01$ / $21.7\pm0.92$ & $0.60\pm0.01$ / $22.9\pm0.77$ \\
    Point Cloud & $\bm{0.64}\pm0.01$ / $\bm{17.5}\pm0.18$ & $\bm{0.65}\pm0.01$ / $\bm{15.6}\pm0.38$ \\
    \bottomrule
    \end{tabular}}
    \vspace*{4mm}
    \includegraphics[width=1\linewidth]{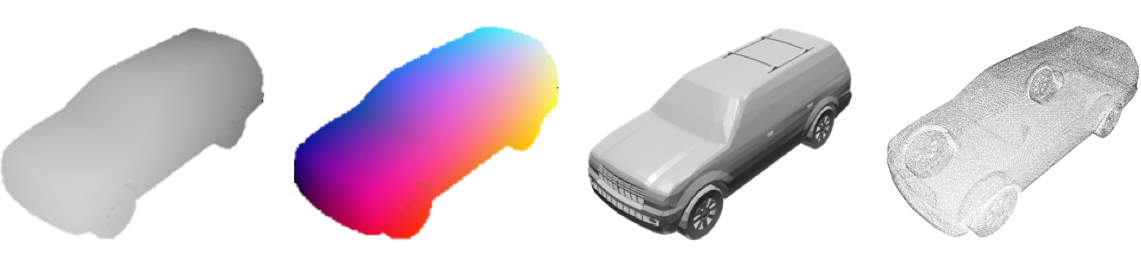}
    \vspace*{-10mm}
\end{table}

We find that using point clouds (with PointNet encoding) provides the best overall performance compared to training with the other 3D representations. This demonstrates the effectiveness of 3D model embedding with point clouds for viewpoint estimation.
By comparing the performance gap between our methods using 3D models, regardless of the choice of 3D representation, and our method without using 3D models (first row in Table~\ref{tab:Ablation3D}), we note again that the 3D models can indeed help improve the viewpoint estimation accuracy on novel classes and reduce the variance introduced by different support training samples.

\myparagraph{Number of exemplars.}
We show detailed evaluation of few-shot viewpoint estimation with different number of shots in Figure~\ref{fig:ViewVaryShot}.
For both the intra-dataset and inter-dataset evaluations, we compute the accuracies and median errors on base and novel classes.
We report the average results and the standard deviations computed over 10 experimental runs with different support training samples.

We first note that all variants of our viewpoint estimation approach can achieve better results when more annotated samples are provided.
Secondly, we find that our approach using only one 3D exemplar model per class clearly improves the performance on both base and novel classes compared to results without using 3D models.
Moreover, adding 3D information also reduces the variance on novel classes, which can clearly be seen in the inter-dataset evaluation.
\revision{This shows that our method without 3D models, which relies on geometrical similarities and consistent labeling between different categories, can already learn a good image embedding space for few-shot viewpoint estimation. Yet, adding 3D models can certainly provide a more direct guidance for a better generalization towards novel categories.}

On the other hand, we note that the performance gap between our approach using a single 3D exemplar per class or using multiple 3D exemplars per class is negligible compared to the gap between using or not 3D models.
\revisionR{It demonstrates that even a single 3D model is sufficient to obtain a good 3D-aware class embedding for viewpoint estimation. It is possible that extra 3D exemplars could prove useful to generate more informative class embeddings, but it would probably require a more sophisticated feature combination than just feature averaging.}

\subsection{Joint Detection and Viewpoint Estimation}
\label{sec:expJoint}

To further demonstrate the generality of our approach in real-world scenarios, we consider the \emph{joint} problem of detecting objects of novel classes in images and estimating their viewpoints.  The fact is that evaluating a viewpoint estimator on ground-truth classes and ground-truth bounding boxes is a toy setting~\cite{starmap2018,Tseng2019FewShotVE}, that is not representative of actual needs.  On the contrary, estimating viewpoints based on predicted detection is much more realistic and challenging. Note that our object detection model and our viewpoint estimation model were trained separately.

\begin{table*}[!t]
\addtolength{\tabcolsep}{-0.5pt}
\centering
\caption{{Evaluation of joint few-shot detection and viewpoint estimation.}
We first recall viewpoint estimation results assuming perfect detection, \ie, using the ground-truth classes and ground-truth bounding boxes (cf.\ Tables \ref{tab:ViewIntra}-\ref{tab:ViewInter}).  Then we use as input predicted classes and estimated bounding boxes given an object detector.  As no code is available to evaluate StarMap+M and MetaView in this setting, we can only evaluate our viewpoint estimation method, for which we used our own detections as input.  (Ours w/o 3D actually does not need to know the class as it is category-agnostic.)
We report the percentage of objects that are correctly detected (right class) with IoU threshold at 0.5, and a rotational error less than $30^\circ$.}
\label{tab:DetView}
\vspace{-1mm}
\scalebox{0.72}{
    \begin{tabular}{l| c c c c c c c c c c c c c c c c c c c c >{\columncolor{HL}}c | c c c c c c c c c c c c >{\columncolor{HL}}c}
    \toprule
    & \multicolumn{21}{c|}{\bf Intra-dataset evaluation on ObjectNet3D} & \multicolumn{13}{c}{\bf Inter-dataset evaluation on Pascal3D+} \\
    Method & \rotatebox[origin=c]{90}{bed} & \rotatebox[origin=c]{90}{bshelf} & \rotatebox[origin=c]{90}{calc} & \rotatebox[origin=c]{90}{cphone} & \rotatebox[origin=c]{90}{comp} & \rotatebox[origin=c]{90}{door} & \rotatebox[origin=c]{90}{fcabin} & \rotatebox[origin=c]{90}{guit} & \rotatebox[origin=c]{90}{iron} & \rotatebox[origin=c]{90}{knife} & \rotatebox[origin=c]{90}{micro} & \rotatebox[origin=c]{90}{pen} & \rotatebox[origin=c]{90}{pot} & \rotatebox[origin=c]{90}{rifle} &
    \rotatebox[origin=c]{90}{shoe} & \rotatebox[origin=c]{90}{slipper} & \rotatebox[origin=c]{90}{stove} & \rotatebox[origin=c]{90}{toilet} & \rotatebox[origin=c]{90}{tub} & \rotatebox[origin=c]{90}{wchair} & \rotatebox[origin=c]{90}{All} 
    & \rotatebox[origin=c]{90}{aero} & \rotatebox[origin=c]{90}{bike} & \rotatebox[origin=c]{90}{boat} & \rotatebox[origin=c]{90}{bottle} & \rotatebox[origin=c]{90}{bus} & \rotatebox[origin=c]{90}{car} & \rotatebox[origin=c]{90}{chair} & \rotatebox[origin=c]{90}{table} & \rotatebox[origin=c]{90}{mbike} & \rotatebox[origin=c]{90}{sofa} & \rotatebox[origin=c]{90}{train} & \rotatebox[origin=c]{90}{tv} & \rotatebox[origin=c]{90}{All} \\ 
    \midrule
    \multicolumn{35}{c}{\bf Evaluated using ground-truth classes and ground-truth bounding boxes (viewpoint estimation)} \\
    \midrule
    StarMap+M\cite{starmap2018} & 32 & 76 & 58 & 59 & 69 & -- & 76 & 59 & \hpz0 & \hpz8 & 82 & -- & 51 & \hpz1 & -- & 15 & 83 & 39 & 41 & 24 & 46
    & \hpz3 & \hpz8 & 11 & 55 & 49 & 21 & 21 & 29 & 11 & 44 & 42 & 64 & 28 \\
    MetaView\cite{Tseng2019FewShotVE} & 36 & 76 & 92 & 58 & 70 & -- & 66 & 63 & 20 & \hpz5 & 77 & -- & 49 & 21 & -- & \hpz7 & 74 & 50 & 29 & 27 & 48
    & 12 & \hpz8 & \hpz9 & 71 & 64 & 22 & 20 & 39 & 14 & 29 & 61 & 58 & 33 \\
    Ours w/o 3D & 53 & 82 & 76 & 54 & 82 & 86 & 83 & 60 & \hpz8 & 21 & 91 & 39 & 64 & 15 & 35 & 19 & 86 & 49 & 50 & 36 & 56 
    & 14 & 30 & 20 & 81 & 73 & 43 & 30 & 51 & 36 & 49 & 62 & 77 & 46 \\
    Ours w/ 3D & 64 & 90 & 90 & 61 & 86 & 90 & 86 & 68 & 34 & 27 & 93 & 47 & 76 & 28 & 49 & 28 & 91 & 69 & 65 & 35 & 65
    & 21 & 33 & 25 & 91 & 74 & 49 & 32 & 59 & 44 & 58 & 72 & 79 & 51 \\
    \midrule
    \multicolumn{35}{c}{\bf Evaluated using predicted classes and predicted bounding boxes (detection + viewpoint estimation)} \\
    \midrule
    Ours w/o 3D & 44 & 73 & 57 & 43 & 48 & 60 & 65 & 60 & \hpz7 & \hpz5 & 55 & 17 & 46 & \hpz4 & 16 & 12 & 76 & 41 & 48 & 19 & 40
    & 14 & 14 & 10 & 12 & 73 & 34 & 19 & \hpz0 & 20 & 41 & 64 & 74 & 31 \\
    Ours w/ 3D & 56 & 75 & 70 & 47 & 53 & 64 & 65 & 75 & 39 & \hpz8 & 57 & 22 & 57 & 15 & 36 & 24 & 82 & 64 & 58 & 24 & 50
    & 15 & 22 & 15 & 15 & 74 & 42 & 16 & \hpz0 & 30 & 54 & 70 & 74 & 35 \\
    \bottomrule
    \end{tabular}}
\end{table*}

\subsubsection{Experimental Setup}

\myparagraph{Datasets.}
As introduced in Section~\ref{sec:expView}, Pascal3D+~\cite{pascal3d14} and ObjectNet3D~\cite{objectnet3d16} are two common viewpoint estimation benchmarks that have already been used in a number of previous publications.
Apart from these two datasets, we also evaluate our method on a more recent benchmark: Pix3D~\cite{sun2018pix3d}. This is a large-scale dataset of 10,069 image-shape pairs with accurate 2D-3D alignment. It contains 395 3D shapes of 9~object categories. Each shape is associated with a set of images capturing the exact object in various environments.

\myparagraph{Evaluation metric.}
As we are considering the joint evaluation of object detection and viewpoint estimation in this section, the metric should reflect the performance of both tasks.
We thus compute the percentage of objects for which the intersection over union between the ground-truth bounding box and the predicted bounding box (with the right class) is larger than 0.5 \emph{and} the rotational error between the ground-truth viewpoint and the predicted viewpoint is smaller than $30^\circ$.
This metric corresponds to the $Acc_{R \frac{\pi}{6}}$ proposed in~\cite{Grabner2019GP2CGP}, which is used to evaluate a joint focal length and 3D pose estimation approach.

\myparagraph{Compared methods.}
We compare our approach to the other viewpoint estimation methods, namely MetaView~\cite{Tseng2019FewShotVE} and StarMap+M, which is the best performing adaptation of StarMap~\cite{starmap2018} (cf.\ Tables \ref{tab:ViewIntra}-\ref{tab:ViewInter}). However, these methods are only evaluated on perfect detections, \ie, ground-truth classes and ground-truth bounding boxes, and no code is available to rerun them on other inputs. Regarding our approach, we consider the case of imperfect detections, where classes and bounding boxes are predicted by our object detector.  Note that the object class is only useful for our viewpoint estimation variant that exploits exemplar 3D models (Ours w/ 3D), as the method variant without 3D information (Ours w/o 3D) is category-agnostic.

\subsubsection{Results}

\myparagraph{Intra-dataset evaluation on ObjectNet3D.}
To experiment with this scenario, we split ObjectNet3D into 80 base classes and 20 novel classes as done in Section~\ref{sec:expView}, and train the object detector and viewpoint estimator using the abundant annotated samples of base classes and scarce labeled samples of novel classes. In this setting, both training and testing samples are from the same dataset, \ie ObjectNet3D.

As recalled in the left part of Table~\ref{tab:DetView}, our few-shot viewpoint estimation outperforms other methods by a large margin when evaluated using ground-truth classes and ground-truth bounding boxes in the 10-shot setting.  When using predicted classes and predicted bounding boxes, accuracy drops for most categories. One explanation is that viewpoint estimation becomes difficult when the objects are truncated by imperfect predicted bounding boxes, especially for tiny objects (\emph{shoes}) and ambiguous objects with similar appearances in different poses (\emph{knives}, \emph{rifles}).
Yet, by comparing the performance gap between, on the one hand, our method when tested using predicted classes and predicted boxes, and, on the other hand, MetaView when tested using ground-truth classes and ground-truth boxes, we find that our approach is able to reach a better accuracy: $50\%$ against 48\%.
This improvement is a strongly encouraging achievement since we free the viewpoint estimation approach from requiring the perfect ground-truth bounding boxes (and classes) without degrading the performance.

\begin{figure*}[!t]
\centering
\includegraphics[width=0.97\linewidth]{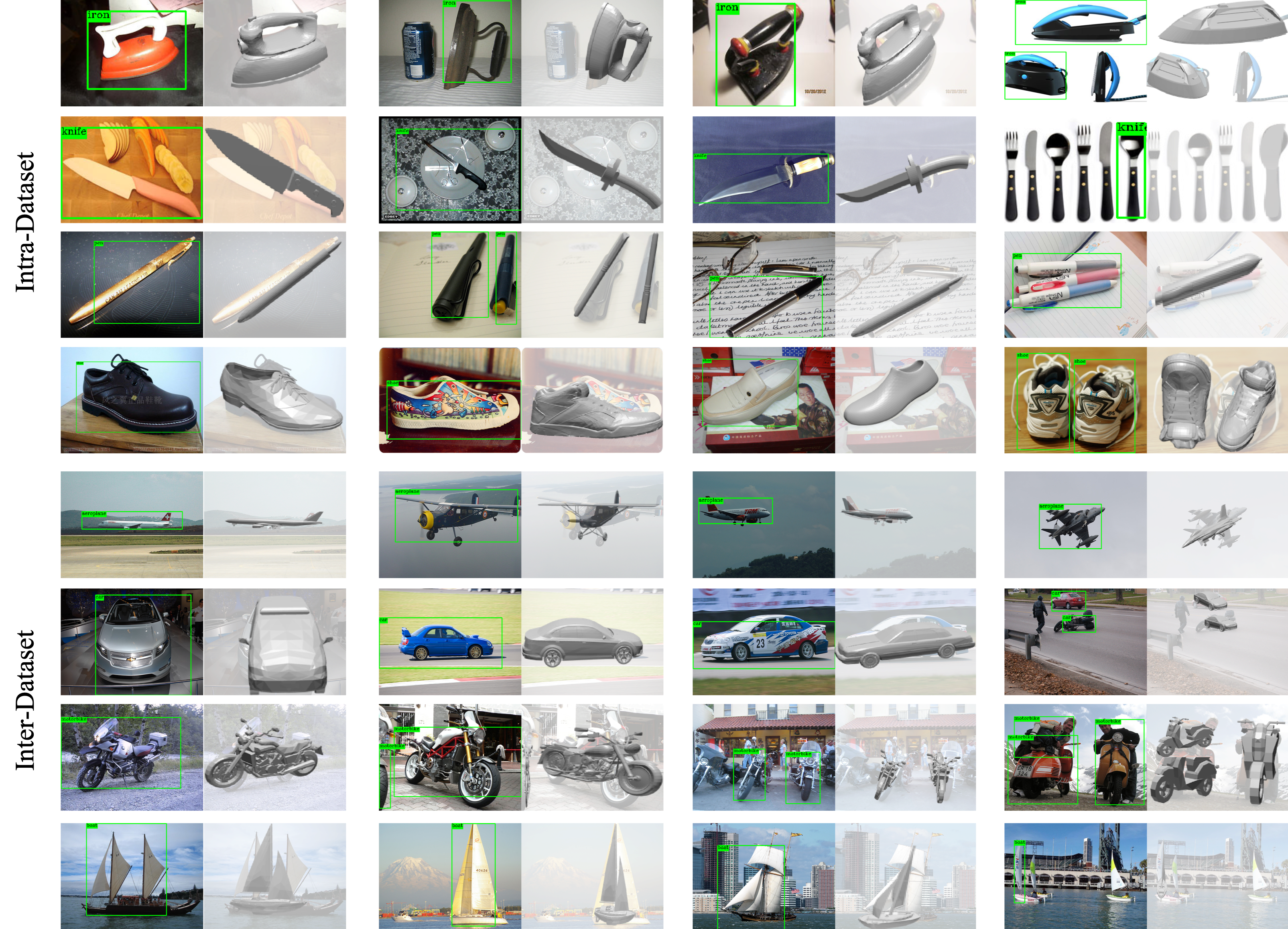}
\vspace{-1mm}
\caption{{Qualitative results of joint few-shot object detection and viewpoint estimation using the predicted 2D bounding boxes given by our object detection model.}
We visualize results on ObjectNet3D and Pascal3D+. For each category, we show three success cases (the first six columns) and one failure case (the last two columns). For each testing image, we project the CAD model of the corresponding class into the predicted 2D bounding box and rotate it according to the estimated viewpoint. 
Error cases include: missing target objects (iron, knife, boat); failed classification (motorbike, car); cluttered objects being detected as one (pen); successful detection but failed viewpoint estimation (shoe and airplane).
}
\label{fig:fsDetView}
\end{figure*}

\myparagraph{Inter-dataset evaluation on Pascal3D+.}
Here, we consider all 12 object categories of Pascal3D+ as novel classes, while the base classes are a set of disjoint object categories from ObjectNet3D and COCO for viewpoint estimation and object detection, respectively.
We use the same split as in the inter-dataset few-shot viewpoint estimation (Section~\ref{sec:expView}), that divides the 100 ObjectNet3D categories into 12 novel ones that intersect with Pascal3D+ and 88 remaining base classes.
Besides, the 12 classes of Pascal3D+ are completely included in the 20 PASCAL VOC object categories, which are set to be the novel classes in the few-shot object detection on MS-COCO (Section~\ref{sec:expDet}).
Therefore, we first use the 10-shot object detection network trained on MS-COCO to detect the novel objects on Pascal3D+, and then, using the predicted 2D bounding boxes, the 10-shot viewpoint estimation network trained on ObjectNet3D.
Unlike the intra-dataset evaluation on ObjectNet3D, our networks are trained and tested on different datasets in this part.

We report the results in the right part of Table~\ref{tab:DetView}. Again, our few-shot viewpoint estimation network outperforms other methods by a large margin when evaluated using ground-truth classes and ground-truth bounding boxes in the 10-shot setting.
Even though a performance drop appears when replacing the ground-truth bounding boxes by the predicted ones, our method using exemplar 3D models still outperforms other methods: 35\% against 33\%. This improvement is especially impressive considering the fact that our object detection and viewpoint estimation networks are both tested on a new dataset that is different from the training datasets, which is a big step towards realistic scenarios and industrial applications.

\begin{table}[!ht]
\addtolength{\tabcolsep}{3pt}
\centering
\caption{Inter-dataset few-shot detection and viewpoint estimation evaluation on Pix3D. $^\dag$Detection network from~\cite{Grabner2019GP2CGP}. $^\ddag$Our few-shot object detection and viewpoint estimation networks trained and tested on different datasets.}
\label{tab:DetViewPix3D}
    \scalebox{0.9}{
    \begin{tabular}{l c c c c >{\columncolor{HL}}c}
    \toprule
    Method & bed & chair & sofa & table & Mean \\
    \midrule
    \multicolumn{6}{c}{\bf Detection + Viewpoint Estimation} \\
    \midrule
    Fine-grained~\cite{wang20183d} & 95 & 88 & 95 & 73 & 88 \\
    GP2C~\cite{Grabner2019GP2CGP} & 98 & 91 & 97 & 77 & 91 \\
    \midrule
    \multicolumn{6}{c}{\bf Detection$^\dag$ + Few-shot Viewpoint Estimation$^\ddag$} \\
    \midrule
    Ours w/o 3D & 81 & 47 & 88 & 53 & 67 \\
    Ours w/ 3D & 86 & 51 & 92 & 58 & 72 \\
    \midrule
    \multicolumn{6}{c}{\bf Few-shot Detection$^\ddag$ + Few-shot Viewpoint Estimation$^\ddag$} \\
    \midrule
    Ours w/o 3D & 68 & 34 & 81 & 13 & 49 \\
    Ours w/ 3D & 71 & 36 & 87 & 14 & 52 \\
    \bottomrule
    \end{tabular}}
\end{table}

\myparagraph{Visual results.}
We provide in Figure~\ref{fig:fsDetView} some qualitative results of few-shot object detection and viewpoint estimation of novel objects on ObjectNet3D and Pascal3D+. For each sample we show the predicted bounding boxes on the left and the estimated viewpoints on the right (visualized by the projected CAD models).
Besides the appearance ambiguities causing major viewpoint estimation errors, we note that the principal failure cases result from the target objects being missed by our object detector (iron and knife) or the objects being wrongly classified (car and motorbike).
Another error is that only one bounding box is predicted for multiple objects of the same class, which usually occurs in cluttered scenes (pen).
These detection errors contribute considerably to the performance drop between evaluating using ground-truth bounding boxes and evaluating using predicted bounding boxes, especially for categories mainly containing tiny objects such as knife in ObjectNet3D and bottle in Pascal3D+.

\myparagraph{Additional results on Pix3D.}
To further demonstrate the effectiveness of our few-shot object detection network and few-shot viewpoint estimation network, we follow GP2C~\cite{Grabner2019GP2CGP} and conduct evaluation on four object categories of Pix3D: \emph{bed}, \emph{chair}, \emph{sofa} and \emph{table}.
As these four classes are completely included in the 12 Pascal3D+ object categories that are considered as novel categories in the inter-dataset evaluation described before, we use the same object detector trained on MS-COCO and viewpoint estimator trained on ObjectNet3D to perform an inter-dataset evaluation on Pix3D.

We first report our results evaluated using the 2D bounding boxes predicted by GP2C in the middle of Table~\ref{tab:DetViewPix3D}. Even though the performance drops from 91\% to 72\%, this result is very encouraging since our viewpoint estimation network has only trained on 10 annotated samples for each testing category while previous methods has trained on thousands of annotated samples. 
Besides using only a small number of annotated training samples of the target classes, our viewpoint estimation network is trained on ObjectNet3D images and directly tested on Pix3D images, while Fine-grained~\cite{wang20183d} and GP2C~\cite{Grabner2019GP2CGP} use images from the same dataset for training and testing. Therefore, our setting is much harder compared to~\cite{wang20183d,Grabner2019GP2CGP}.
We then report our results evaluated using predicted bounding boxes given by our few-shot object detector at the bottom of Table~\ref{tab:DetViewPix3D}.
The overall performance drops around 20\% points compared to the evaluation using bounding boxes predicted by GP2C, where the detection network is pre-trained on all 80 object categories of MS-COCO and fine-tuned on the 4 categories of Pix3D.
In both cases, our viewpoint estimation method using 3D models performs better than our method without 3D models. This consistent improvement demonstrates again the benefits of adding 3D information in viewpoint estimation.

\revisionR{
\subsection{Limitations}\label{sec:limitations}
Our work shares a common limitation with other work on viewpoint estimation in that it does not handle very well small objects, which have less visible cues, and objects that are nearly symmetrical, such as \emph{knives}. In the latter case, a wrong prediction of the front-back orientation can result in a very large prediction error, although the rendered views can be very similar to the actual images. 
It is even more so in the few-shot setting, where only a few labeled samples are provided for the novel categories. 
Preventing such failure cases could require a specific treatment of almost-symmetries.}

\revision{Also, as discussed in the introduction regarding the case where we do not use 3D model information for viewpoint estimation but a class-agnostic approach instead, we rely on the fact that objects of different but related classes often are consistently oriented, with aligned similarities. While it is the case for all datasets we know of, this fact is not guaranteed. Yet, in case of orientation discrepancies between classes, a dataset can somehow be ``normalized'' before training by applying systematic rotations of ground-truth viewpoints.}


\section{Conclusion and Perspectives}

In this work, we presented an approach to few-shot object detection and viewpoint estimation that can tackle both tasks in a coherent and efficient framework.
We demonstrated the benefits of this approach in terms of accuracy, and significantly improved the state of the art on several standard benchmarks for few-shot object detection and few-shot viewpoint estimation.
Moreover, we showed that our few-shot viewpoint estimation model can achieve promising results on the novel objects detected by our few-shot detection model, compared in an adversarial setting to other existing methods tested on perfect detection, \ie, ground-truth classes and ground-truth bounding boxes.

This is of particular interest for scene understanding in weakly-controlled environments, such as robotic manipulation with various objects in the wild. In future work, we are interested in developing category-agnostic models that can detect arbitrary objects and estimate their poses without seeing them during training.
We will also expand our approach to perform 3D model retrieval and estimation refinement by selecting the 3D candidate that best agrees with the measured visual evidence, which might include RGB images, depth maps, and deep features extracted by a neural network.
The exploitation of multiple views and additional inputs such as depth maps could also be considered.



\ifCLASSOPTIONcaptionsoff
  \newpage
\fi

\bibliographystyle{IEEEtran}
\bibliography{bibtex/bib/IEEEabrv.bib,bibtex/bib/IEEEexample.bib}{}

\begin{IEEEbiography}[{\includegraphics[width=1in,height=1.25in,clip,keepaspectratio]{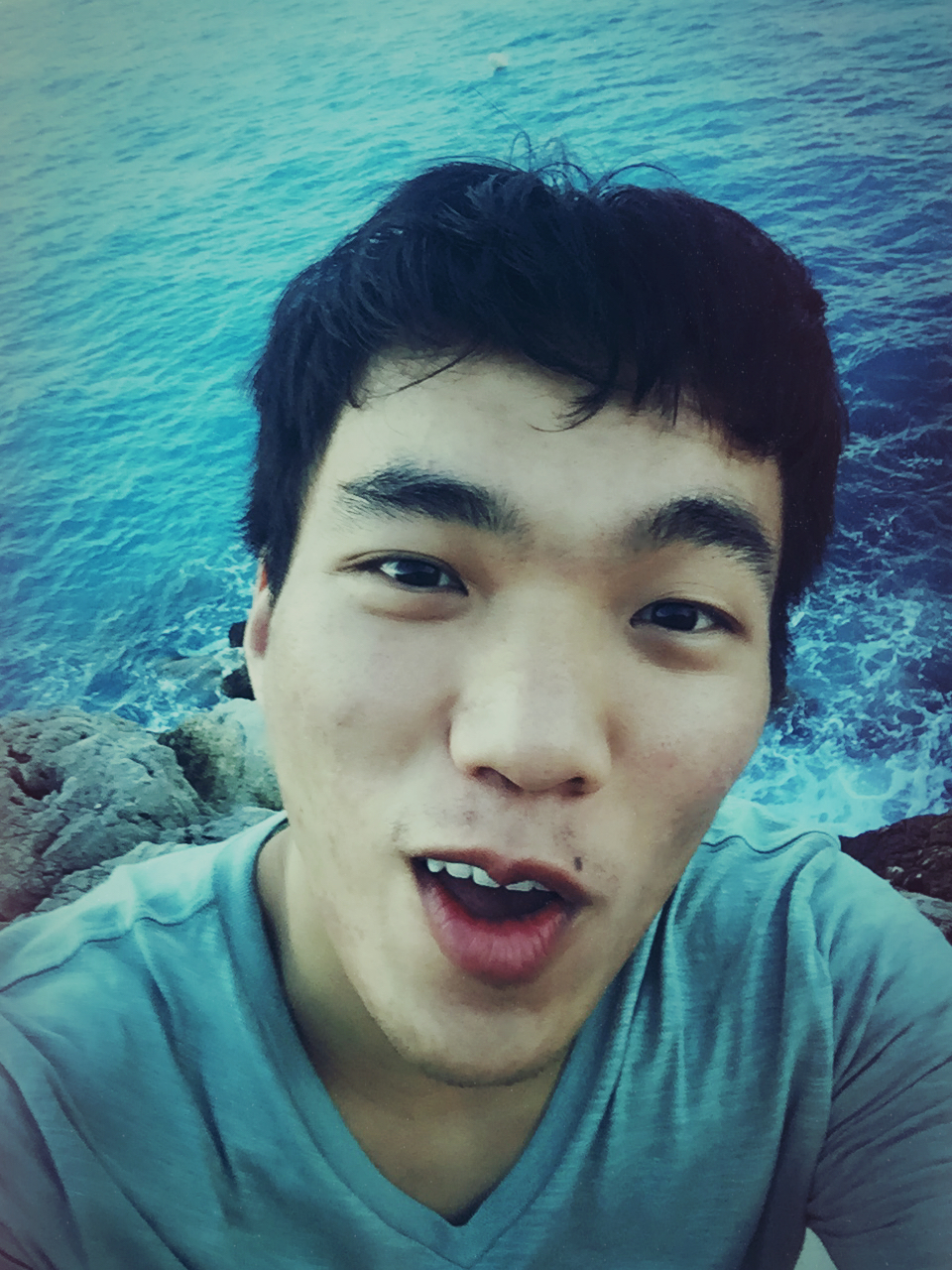}}]{Yang Xiao} is a Ph.D. candidate in Computer Science at Ecole des Ponts ParisTech, France. 
He received his M.S. in Signal and Image Processing from University Paris-Saclay and B.S. in Optical and Electronic Information from Huazhong University of Science and Technology.
His research interests include object pose estimation and 3D scene understanding in computer vision. 
\end{IEEEbiography}

\begin{IEEEbiography}[{\includegraphics[width=1in,height=1.25in,clip,keepaspectratio]{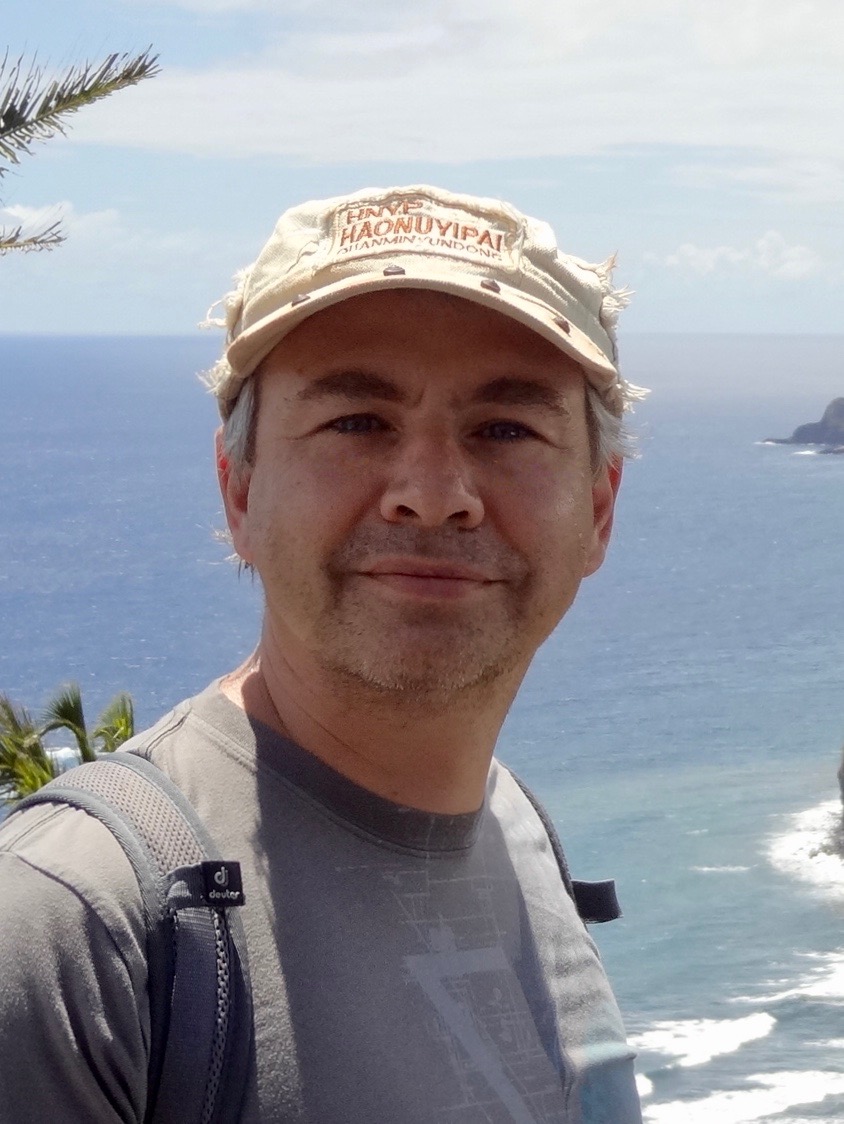}}]{Vincent Lepetit} is a 
research director at ENPC ParisTech, France. Before that, he was a full professor at the Institute for Computer Graphics and Vision, Graz University of Technology, and before that, a senior researcher at CVLab, EPFL, Switzerland.
He currently focuses on 3D scene understanding from images, with application to 3D hand and object tracking, 3D reconstruction, and camera localization. With his co-authors, he received the Koenderick 'test of time' award for the BRIEF local descriptor in 2020.
\end{IEEEbiography}

\begin{IEEEbiography}[{\includegraphics[width=1in,height=1.25in,clip,keepaspectratio]{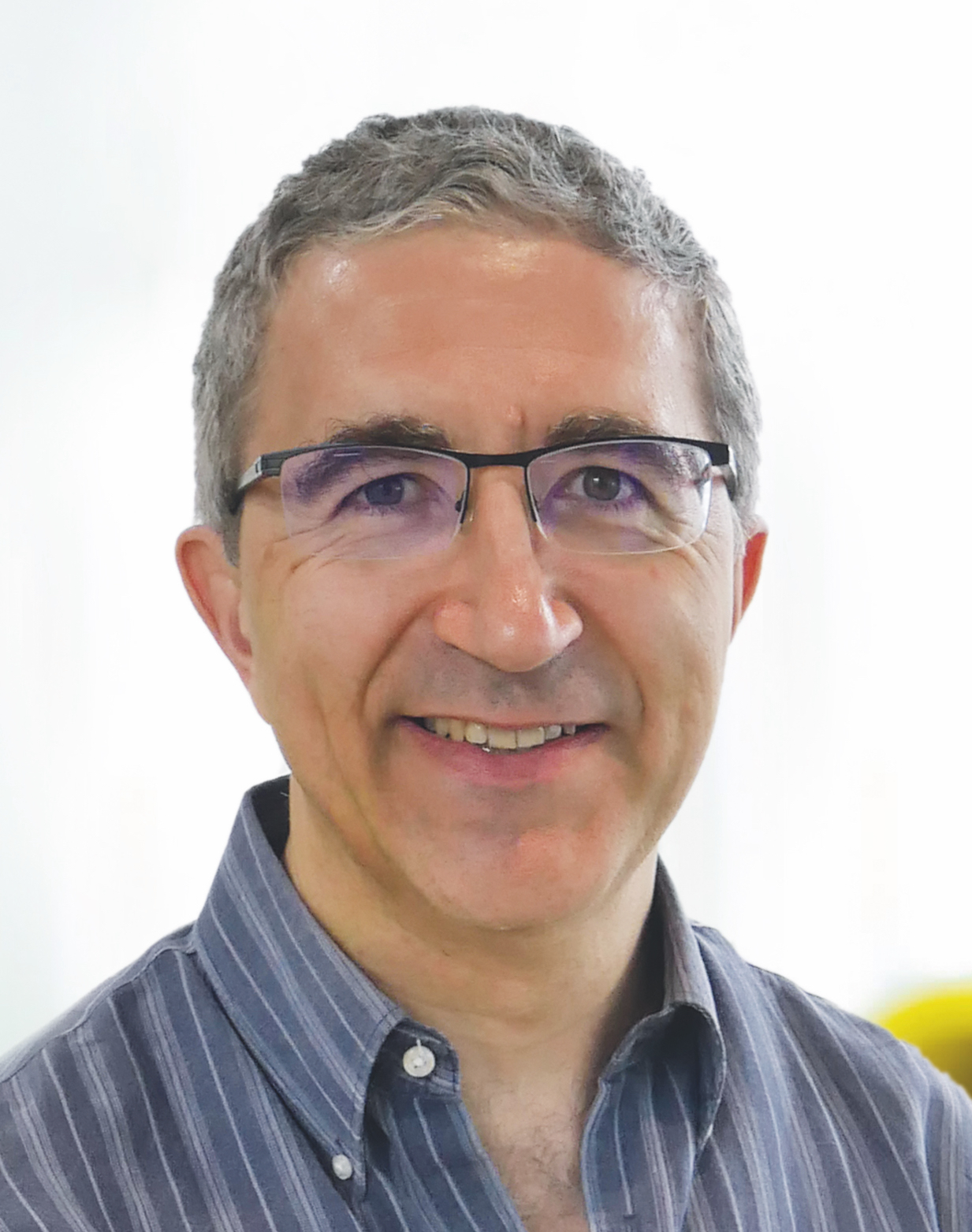}}]{Renaud Marlet} is a Senior Researcher at \'Ecole des Ponts ParisTech (ENPC) and a Principal Scientist at valeo.ai, France. He has held positions both in academia (researcher at Inria) and in the software industry (expert at Simulog, deputy CTO of Trusted Logic). He was the head of the IMAGINE group at LIGM/ENPC (2010-2019). He is currently interested in scene understanding and semantized 3D reconstruction, with applications to robotics, autonomous driving and civil engineering.
\end{IEEEbiography}


\end{document}